\documentclass{article}

\usepackage[final]{neurips_2025}

\usepackage[utf8]{inputenc} 
\usepackage[T1]{fontenc}    
\usepackage{hyperref}       
\usepackage{url}            
\usepackage{booktabs}       
\usepackage{amsfonts}       
\usepackage{nicefrac}       
\usepackage{microtype}      
\usepackage{xcolor}         

\usepackage{microtype}
\usepackage{graphicx}
\usepackage{booktabs} 
\usepackage{multirow}
\usepackage{hyperref}
\usepackage{enumitem}
\usepackage[T1]{fontenc}
\usepackage{xspace}

\usepackage{amsmath}
\usepackage{amssymb}
\usepackage{mathtools}
\usepackage{amsthm}
\usepackage{lscape}
\usepackage{rotating}
\usepackage{xurl}
\usepackage{subcaption}
\usepackage{caption}
\usepackage{todonotes}
\usepackage{wrapfig}

\usepackage{authblk}

\newcommand\ourframework{\textcolor{black}{CATransformers}\xspace}

\title{CATransformers: Carbon Aware Transformers Through Joint Model-Hardware Optimization}

\author{%
  Irene Wang$^{1,2,*}$, Mostafa Elhoushi$^{2}$, H. Ekin Sumbul$^{3}$, Samuel Hsia$^{2}$, Daniel Jiang$^{4}$, Newsha Ardalani$^{2}$, Divya Mahajan$^{1}$, Carole-Jean Wu$^{2}$, Bilge Acun$^{2}$\\
  $^{1}$Georgia Institute of Technology, $^{2}$FAIR at Meta, $^{3}$Reality Labs at Meta, $^{4}$Meta,  \\
  $^{*}$Work done at Meta \\
  \texttt{irene.wang@gatech.edu} \hspace{2ex} \texttt{acun@meta.com}
}

\begin{document}

\maketitle

\begin{abstract}

Machine learning solutions are rapidly adopted to enable a variety of key use cases, from conversational AI assistants to scientific discovery. This growing adoption is expected to increase the associated lifecycle carbon footprint, including both \emph{operational carbon} from training and inference and \emph{embodied carbon} from AI hardware manufacturing. 
We introduce \ourframework---the first carbon-aware co-optimization framework for Transformer-based models and hardware accelerators. By integrating both operational and embodied carbon into early-stage design space exploration, \ourframework enables sustainability-driven model architecture and hardware accelerator co-design that reveals fundamentally different trade-offs than latency- or energy-centric approaches. Evaluated across a range of Transformer models, \ourframework consistently demonstrates the potential to reduce total carbon emissions---by up to 30\%---while maintaining accuracy and latency. We further highlight its extensibility through a focused case study on multi-modal models. Our results emphasize the need for holistic optimization methods that prioritize carbon efficiency without compromising model capability and execution time performance. The source code of \ourframework is available at {\small{\href{https://github.com/facebookresearch/CATransformers}{\texttt{https://github.com/facebookresearch/CATransformers}}}}.

\end{abstract}
\section{Introduction}


As machine learning (ML) systems become more widespread across various industries, it is crucial to take a closer examination of their carbon footprint and find strategies to mitigate it across the system stack. 
%
%
Sustainable ML system design requires a holistic approach that considers both operational carbon (energy used during training and inference) and embodied carbon (emissions from hardware manufacturing and lifecycle) \cite{act}, which together contribute to total emissions.
This work tackles the question: \textit{How does incorporating carbon footprint metrics into optimization workflows influence the design of ML models and hardware architectures as a co-optimization?} 

A fundamental challenge in designing sustainable AI systems stem from the tight coupling between model architecture and hardware design, which together shape both operational and embodied carbon. Model execution on a particular hardware determines runtime characteristics, such as compute intensity and memory access patterns, which affect operational carbon, while hardware parameters like chip area, memory hierarchy, and fabrication technology drive embodied carbon. Carbon optimization requires joint model-hardware exploration, particularly during early accelerator design, to uncover opportunities that align model demands with hardware capabilities in a carbon-efficient way. Co-design is even more important for multi-modal workloads like vision-language models. Each modality introduces distinct bottlenecks—e.g., vision transformers usually have compute-bound workloads~\cite{modelquant}, while text transformers have more memory-bound workloads~\cite{longtext}.
%
Co-optimization across modalities and system layers is essential to minimize the total carbon footprint without compromising performance.

While this sustainability challenge applies across the entire AI lifecycle, the rapid adoption of AI on edge devices presents a particularly acute and fast-growing problem~\cite{FLcarbon, chasingCarbon}. With billions of products like smartphones and AR/VR headsets requiring specialized, custom accelerators~\cite{3d, edge_avatars, hailo, edgetpu, jetson-orin}, the embodied carbon from manufacturing constitutes a massive, upfront environmental cost. Furthermore, the cumulative operational carbon from continuous inference over a device's multi-year lifespan represents a substantial and often overlooked energy demand~\cite{apple2021}. Addressing the tightly coupled model-hardware system at the edge is therefore a critical first step, offering a high-impact domain to develop and validate the carbon-aware co-design principles needed for sustainable AI.

Prior hardware-aware neural architecture search (NAS) methods have focused primarily on latency or energy and are not suitable for total carbon optimization. This is because carbon is a complex metric that cannot be captured by optimizing latency or energy in isolation. These approaches either neglect embodied carbon by optimizing only the model on fixed hardware~\cite{hat} or treat latency, energy, and chip area as orthogonal constraints~\cite{naas}, overlooking the tightly coupled nature of operational (latency- and energy-driven) and embodied (chip area-driven) carbon. As a result, these methods are fundamentally limited in their ability to minimize total carbon emissions. 
Furthermore, prior work often assumes fixed hardware or fixed model backbones, limiting their applicability to a broader range of emerging hardware platforms and modern,  multi-modal models.

We introduce \ourframework---\textit{the first framework to jointly optimize model and hardware architecture with the specific goal of minimizing total carbon emissions for edge inference-only devices.} Distinct from prior latency- or energy-centric methods, our approach incorporates both operational and embodied carbon into a \textbf{unified optimization pipeline}, revealing carbon-efficient configurations previously overlooked. We observe that optimizing traditional metrics in isolation frequently results in suboptimal trade-offs. For example, accelerating inference by increasing compute density may inadvertently increase power consumption and embodied carbon due to larger chip area or more resource-intensive fabrication. By prioritizing total carbon as a first-class objective, \ourframework yields fundamentally different and more sustainable design choices.


\ourframework\ systematically explores the joint design space of model architectures and hardware accelerators using multi-objective optimization. It comprises three main components: (1) a \textit{Multi-Objective Bayesian Optimizer} that balances accuracy, latency, energy, and carbon emissions; (2) an \textit{ML Model Evaluator} that efficiently navigates model variants using importance-based pruning and fine-tuning; and (3) a \textit{Hardware Estimator} that profiles latency, energy, and carbon footprint.

We evaluate \ourframework\ on language, vision, and multi-modal Transformer-based models and show that its co-optimized model–hardware configurations reduce total carbon by 30\% over latency-optimized and 8\% over energy-optimized baselines. To further highlight its utility, we present CarbonCLIP, a family of \ourframework-optimized CLIP models. 
%
%
\begin{figure} [t]
\centering
\vspace{-4ex}
\includegraphics[width=0.6\textwidth]{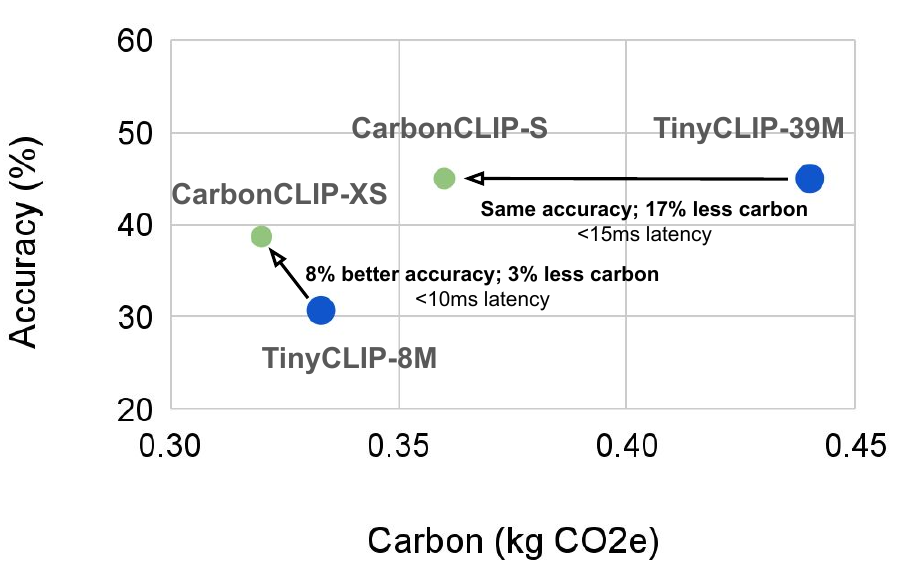}
\vspace{-1ex}
\caption{CarbonCLIP models achieve lower carbon footprint and higher accuracy compared to baseline CLIP models.}
\label{fig:accuracy}
\vspace{-2ex}
\end{figure}

Figure~\ref{fig:accuracy} shows CarbonCLIP achieves up to 17\% lower total carbon emissions compared to state-of-the-art CLIP variants on edge devices, while maintaining similar accuracy and latency. Joint optimization enables low-carbon, high-accuracy designs that outperform approaches that optimize only hardware configurations.


    
Our key contributions include:
\begin{enumerate}
    \item \textbf{Insights and Analysis:} Empirical insights showing how carbon-aware optimization shifts model–hardware trade-offs relative to traditional metrics, enabling early-stage design exploration for next-generation ML accelerators.
    
   \item \textbf{Quantification Framework}:  We develop the first open-source toolchain to estimate both operational and embodied carbon for custom accelerators during design-space exploration.

   \item \textbf{Carbon-Aware Co-optimization:} \ourframework\ uses multi-objective Bayesian optimization to jointly explore model and hardware design spaces, balancing accuracy, latency, energy, and carbon. It leverages fine-tuning–based proxy signals to reduce evaluation cost.

   \item \textbf{Sustainable Multi-Modal Models}: Using \ourframework, we co-optimize CLIP variants (CarbonCLIP) that reduce total carbon by up to 17\% compared to edge-deployed CLIP baselines, without sacrificing accuracy or latency. 
\end{enumerate}


%
 \ourframework is designed as a \textbf{modular, extensible framework} that supports the seamless integration of diverse model architectures, optimization strategies, and carbon estimation techniques. The goal is not to outperform other NAS frameworks on traditional metrics, but to highlight the fundamentally different design trade-offs that emerge under carbon-centric optimization objectives. Through systematic comparisons across multiple optimization modes and model families, we demonstrate the practical impact of carbon-aware design, which is the core contribution of this work.

\section{Background and Related Works}
\label{sec:motivation}

\textbf{Hardware Accelerator Search:}
Specialized hardware accelerators have been developed to efficiently run deep learning workloads, with tensor cores for matrix operations~\cite{tpuv4_isca,tabla,cosmic:micro, chen2016eyeriss} and vector cores for element-wise operations~\cite{tpu, tandem:asplos:2024}. Prior accelerator search frameworks~\cite{wham, wham-patent, phaze, phaze-patent, fast, spotlight} optimize for throughput and energy, but do not jointly optimize hardware and model architectures or consider carbon footprint in the design process.

\textbf{Hardware and Neural Architecture Co-optimization:}
Co-optimization methods~\cite{naas, dance, coop, naas_2} typically focus on latency or energy, ignoring carbon as a primary design objective. Approaches that do consider carbon~\cite{elgamal2023carbon, CORDOBA, carbonNas} often optimize either the model or hardware in isolation. Furthermore, prior work often assumes fixed hardware or fixed model backbones, limiting applicability to a broader range of emerging hardware platforms and modern, heterogeneous models—particularly multi-modal architectures. Our work addresses this gap by jointly optimizing both hardware and model architectures with total carbon footprint—including embodied and operational emissions—as a first-class objective.

\textbf{The Carbon Footprint of AI Systems:} Much of the research on the carbon footprint of AI focuses on operational emissions, with three main directions: (1) quantifying the emissions from training~\cite{carbon_1, carbon_5, mlco2, sustainableai}, (2) analyzing energy use during deployment and inference~\cite{carbon_2, carbon_3}, and (3) developing techniques to reduce emissions while exploring energy–performance trade-offs~\cite{carbon_4, cenas, sprout, carbondep}.

Embodied carbon, however, remains underexplored. Tools like ACT \cite{act}, IMEC.netzero \cite{imec}, and LLMCarbon~\cite{llmcarbon} have begun to estimate embodied emissions, but a holistic framework that quantifies and minimizes both operational and embodied carbon remains lacking. CORDOBA~\cite{CORDOBA} introduces a carbon-aware accelerator design tool, but it is not open-sourced, limiting reproducibility and extensibility. Unlike CE-NAS~\cite{cenas}, which reduces carbon emissions during neural architecture search, \ourframework is the first to optimize total carbon emissions from both model and hardware, during early-stage design.

\textbf{AI Systems on the Edge:}
AI models are no longer confined to large-scale datacenters and are now widely deployed on edge devices like smartphones, AR/VR headsets, robotics, and vehicles~\cite{FLcarbon, edge_avatars, hailo}. These edge settings present unique challenges, including strict resource constraints, hardware heterogeneity, and communication bottlenecks that impact performance~\cite{fluid, glueFL, edgeImpulse}. This wide adoption, however, also raises significant environmental concerns. The sheer scale of device production means embodied carbon from chip fabrication is a major contributor to total emissions~\cite{chasingCarbon}. Simultaneously, the cumulative operational carbon from frequent inference over a device's multi-year lifespan presents a distinct and growing sustainability challenge. \ourframework aims to address this gap by enabling holistic, carbon-aware co-design of AI models and hardware accelerators specifically for edge systems.

\textbf{CLIP Models and Edge Variants:}
Multi-modal models like CLIP~\cite{clip} combine Transformer-based text encoders with image encoders (ResNet~\cite{resnet} or ViT~\cite{vit}) and are trained on large datasets~\cite{laion400m, metaclip, laion5b, datacomp} to learn cross-modal associations. These models support tasks like zero-shot classification and retrieval. Several variants~\cite{tinyclip, mopeclip, upop, mobileclip} adapt CLIP for edge devices through pruning and efficient training, but focus solely on accuracy and latency. Our work is the first to optimize CLIP models for edge deployment with total carbon emissions in mind, without sacrificing accuracy or latency.

\begin{figure*}[t]
\centering
\includegraphics[width=0.9\textwidth]{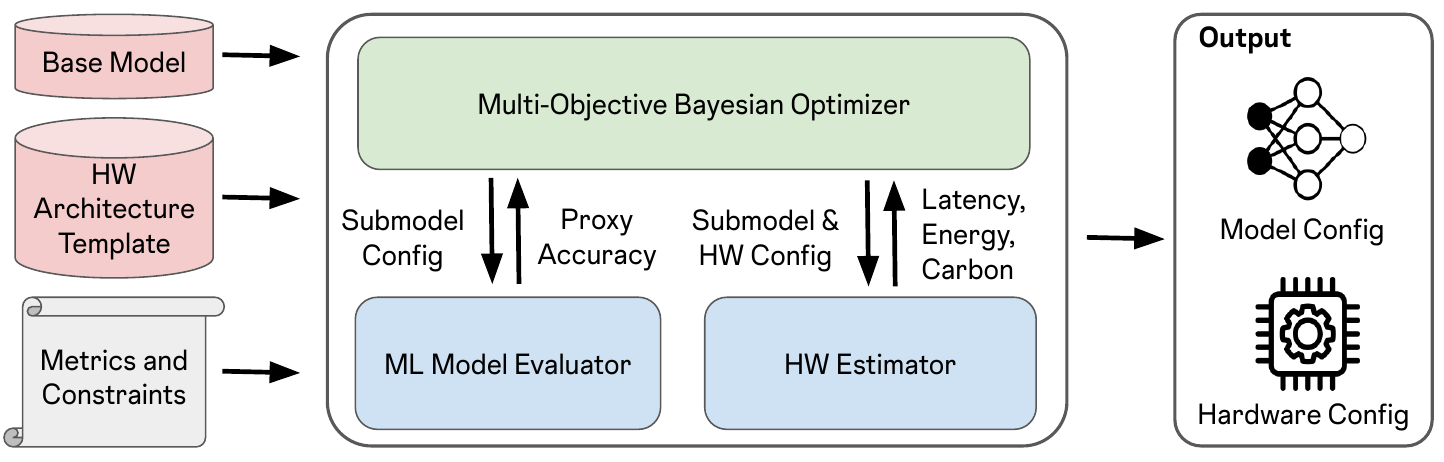}
\vspace{-1ex}
\caption{Overview of the \ourframework framework. The Bayesian optimizer iteratively explores model and hardware configurations using accuracy, carbon, and latency estimates from evaluation modules, outputting optimized co-designs.}
\label{fig:overview}
\vspace{-1ex}
\end{figure*} 

\section{Framework Overview}
\label{sec:overview}

In this section, we introduce \ourframework, a carbon-aware architecture search framework for sustainability-driven co-optimization of ML models and hardware architectures.
As illustrated in Figure~\ref{fig:overview}, \ourframework framework takes three inputs: (1) a base ML model, (2) a hardware architecture template, and (3) a set of optimization objectives and constraints that define the joint model–hardware search space. The framework consists of three key components: a multi-objective optimizer, an ML model evaluator, and a hardware estimator. 

\subsection{\ourframework Inputs}

\begin{figure}[t]
    \centering
    \begin{subfigure}[t]{0.47\textwidth}
        \centering
        \includegraphics[trim={0 0 0 0.7cm},clip, width=0.6\textwidth]{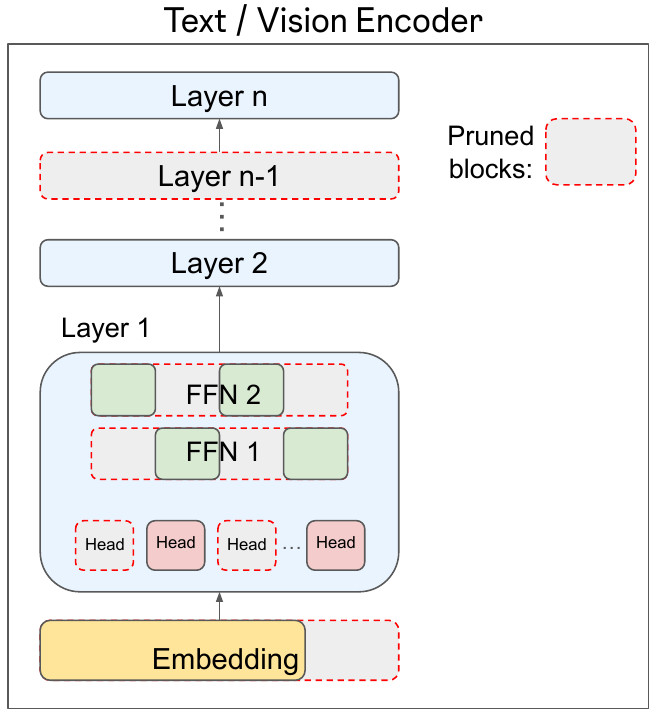}
        \caption{Overview of the pruning dimensions}
        \label{fig:pruning}
    \end{subfigure}
    \hfill
    \begin{subfigure}[t]{0.45\textwidth}
        \centering
        \includegraphics[width=\textwidth]{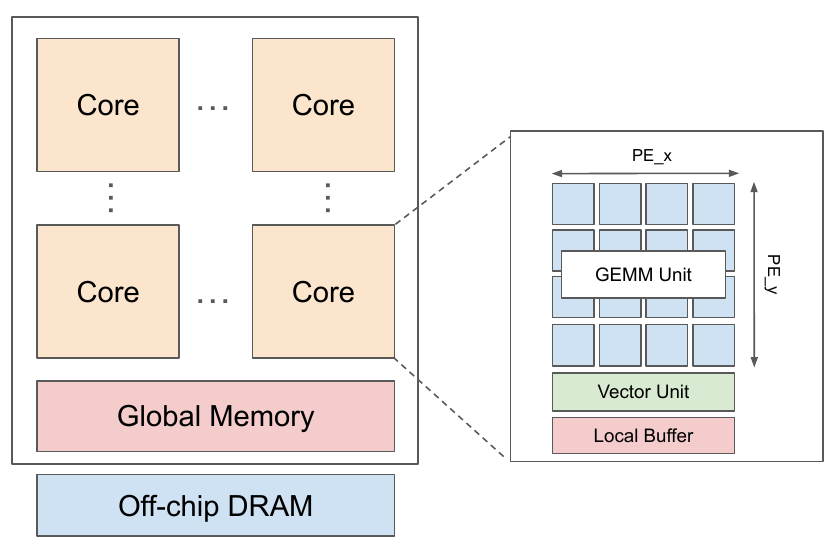}
        \caption{Architecture template}
        \label{fig:template}
    \end{subfigure}
    \caption{Overview of model pruning dimensions and hardware template for \ourframework }
    \vspace{-1ex}
\end{figure}

\textbf{Base Model:} The base model is a large, pre-trained Transformer that serves as the foundation for generating pruned variants. It defines the architecture's shape, functionality, and search space. As shown in Figure~\ref{fig:pruning}, pruning is applied along several dimensions, including the number of layers, hidden size, feedforward network width, attention heads, and embedding dimensions.

\textbf{Hardware Architecture Template:} 
The accelerator template (Figure~\ref{fig:template}) captures key components and tunable parameters inspired by academic and industry designs~\cite{tpu, wham, fast, phaze}. It features tensor cores with Processing Elements (PEs) arranged in $X$ and $Y$ dimensions for GEMM operations, vector units for element-wise operations, local buffers for data reuse, and a shared global SRAM and off-chip memory—consistent with existing edge inference accelerators~\cite{edge_avatars, 3d}. Increasing PEs or cores boosts performance but raises area and energy costs; more on-chip memory lowers latency but adds energy overhead. The optimal hardware configuration depends on the model architecture, size,
and performance constraints. Table~\ref{tab:design_space} summarizes the hardware design space parameters.

\begin{scriptsize}
\begin{table}[t]
\centering
\vspace{-1.5ex}
\caption{Architecture design space parameters.}
\vspace{1ex}
\resizebox{0.5\columnwidth}{!}{
\begin{tabular}{ l c l }
 \hline
 \textbf{Parameter Description} & \textbf{Notation} & \textbf{Potential Values} \\
 \hline
 \multicolumn{3}{l}{\textbf{Design Space ($S$)}} \\
 \hline
 Number of Cores & TC & 1 to 4 powers of 2 \\
 PE Array X dim & $\mathrm{PE}_x$ & 1 to 256 powers of 2 \\
 PE Array Y dim & $\mathrm{PE}_y$ & 1 to 256 powers of 2 \\
 Global Buffer Size & GLB & 1 to 8~MB powers of 2 \\
 Local Buffer Size & $\mathrm{L2}$ & 256~KB to 4~MB powers of 2 \\
 Local Bandwidth & $\mathrm{L2}_{\mathrm{bw}}$ & 1 to 256 words/cycle \\
 Vector Unit width & $V_{\mathrm{pe}}$ & $=\mathrm{PE}_x$ \\
 \hline
 \multicolumn{3}{l}{\textbf{Fixed Parameters}} \\ 
 \hline
 Global Bandwidth & GLB-BW & 256 words/cycle \\
 Off-chip DRAM Size & HBM & 1~GB \\
 Technology & Tech & 22~nm \\
 BitWidth & $B$ & 8 \\ 
 Maximum TOPS & $T_\mathrm{max}$ & 20 TOPS \\
 Frequency & $f$ & 500 MHz \\
 \hline
\end{tabular}}
\label{tab:design_space}
\vspace{-1.5ex}
\end{table}
\end{scriptsize}

\subsection{ML Model Evaluator}
\label{sec:ml_model}
The ML model evaluator estimates the accuracy of candidate model architectures during the search.

\textbf{Pruning}: The evaluator prunes the pre-trained base model along key dimensions, including the number of layers, feedforward network size, attention heads, and embedding dimension. For multi-modal models like CLIP, each Transformer (text and vision) is pruned independently using strategies from prior work~\cite{mopeclip, poormanbert, hat}. Each layer is pruned uniformly. See Appendix~\ref{app:pruning} for pruning details and empirical observations.

\textbf{Fine-tuning for Accuracy Proxy}: 
Direct evaluation of untrained pruned models results in poor accuracy, while fully retraining every model is computationally infeasible. Instead, we use lightweight fine-tuning to approximate accuracy. Our ablation studies show that this method yields a high Spearman correlation (0.98) with fully trained accuracy, making it a reliable proxy for ranking candidate models (Appendix~\ref{app:proxy}).
This approach allows accurate and efficient evaluation within the optimization loop, enabling scalable exploration of Transformer-based models without full retraining.
%
%
For language models we use the MRPC task from the GLUE dataset~\cite{glue} for semantic similarity detection, for vision models, the CIFAR-10 dataset~\cite{cifar10} for image classification; and for CLIP models, the MSCOCO dataset~\cite{mscoco} for retrieval.




\subsection{Hardware Estimator}
\label{sec:estimator}

The hardware estimator provides unified, end-to-end analysis of inference latency and total carbon footprint—including both embodied and operational emissions—for each model-hardware configuration. It leverages performance and energy models to estimate operator-level latency, energy, and area~\cite{accelergy, sunstone}, and incorporates carbon modeling for manufacturing-related emissions and location-specific operational carbon, scaled over the system’s deployment lifetime~\cite{act,electricitymaps}.
To the best of our knowledge, this represents the first open-source toolchain built for early-stage design space exploration that holistically quantifies latency, energy, and total carbon emissions of custom ML accelerators. A detailed workflow is provided in Appendix~\ref{app:estimation_tools}.

\subsection{Multi-Objective Optimization}

\ourframework uses multi-objective Bayesian optimization to efficiently explore the joint design space of model architectures and hardware configurations building on Ax~\cite{Ax} and BoTorch~\cite{balandat2020botorch} with the qNEHVI algorithm~\cite{qNEHVI}. Compared to reinforcement learning or evolutionary search, Bayesian optimization offers better sample efficiency and uncertainty modeling, making it ideal for our static, high-dimensional search space where each configuration takes 5-15 minutes to evaluate. While prior work~\cite{hat} explored just 125 models over 30 iterations, \ourframework scales to a search space of ~100 million configurations in only 100 iterations.

The optimization maximizes accuracy while minimizing latency, energy, and total carbon (embodied + operational), exploring model (Figure~\ref{fig:pruning}) and hardware parameters (Table~\ref{tab:design_space}) under a TOPS-based compute budget based on publicly available edge accelerators. Our framework supports four compute-constrained modes: (1) Accuracy \& Total Carbon (with latency constraint), (2) Accuracy \& Latency, (3) Accuracy \& Energy, and (4) Accuracy, Latency, \& Total Carbon.


\subsection{Outputs}
\ourframework outputs optimized combinations of model and hardware configurations that, when deployed together, improve overall efficiency. Once the optimizer identifies carbon-efficient, pruned model architectures, these models can optionally be fine-tuned to recover any accuracy loss. 
Preliminary results show these pruned models require far fewer training steps than full pre-training. Specifically, CarbonCLIP models are fine-tuned for just 2 epochs on MetaCLIP~\cite{metaclip}, using only 40\% of the training steps compared to prior works~\cite{clip, metaclip, tinyclip}.

\section{Evaluation}
\label{sec:eval}
\subsection{Experimental Settings}
\label{sec:setup}
\textbf{Model:} 
To demonstrate the versatility of \ourframework, we apply it to diverse transformer-based architectures: encoder-only Bert-Base~\cite{bert}, decoder-only Llama3-8B~\cite{llama3}, vision transformer ViT-B/16~\cite{vit}, and multi-modal CLIP models (CLIP-ViT-B/16 and CLIP-ViT-B/32)\cite{clip}. We use pre-trained weights from HuggingFace\cite{huggingFace} for Bert, Llama, and ViT, and adopt OpenCLIP’s~\cite{openclip} DataComp-1B~\cite{datacomp} models for CLIP. Building on CLIP optimization results, we generate \textbf{CarbonCLIP} models with corresponding hardware designs.

\textbf{Baselines:} Prior work primarily targets performance or energy efficiency, rarely considering carbon-aware hardware-model co-design. These approaches are typically evaluated on fixed hardware and model architectures, making direct comparison difficult. To ensure a fair and meaningful evaluation, we benchmark CarbonCLIP against two widely adopted baselines: (1) standard CLIP models (ViT-B/16, ViT-B/32), and (2) TinyCLIP, a state-of-the-art variant optimized for latency. For each, we apply \ourframework's hardware search with fixed model architectures to derive comparable accelerator designs. This allows us to isolate the benefits of carbon-aware co-optimization. Our results show that carbon-optimized models can match or exceed the accuracy and latency of traditional baselines—demonstrating the practical value of our approach.

\textbf{Hardware Estimation and Validation:}
\label{sec:val}
We estimate accelerator area, latency, and energy using open-source tools (Appendix~\ref{app:estimation_tools}), assuming 22nm process technology and 8-bit integer operations.
To validate latency estimates, we use SCALE-Sim~\cite{scalesimv1,scalesimv2} for CLIP’s QKV projections. Our estimates have an average latency error of 13\%, consistent with prior work~\cite{timeloop,accelergy}, and are validated on standard compute array sizes (16×16 to 64×64).
We further validate energy and latency predictions against real GPU hardware (V100, A100, H100) by modeling GPU-like architectures. Our estimates closely match measured results, with average errors of 8\% (energy) and 9\% (latency), and Spearman’s rank-order correlation $r$ ranging from 0.5 to 1.0 and $p < 0.05$, confirming toolchain accuracy. Details of the validation is in Appendix~\ref{app:gpu_validation}.

\textbf{Carbon Emission Estimation:}
Operational carbon is estimated using California grid intensity over a 3-year lifespan (1 inference/sec, 6 hrs/day), reflecting typical mobile usage~\cite{apple2021,nsys2025,harmonyhit2025}. Embodied carbon is calculated assuming fabrication in Taiwan. Appendix~\ref{app:region} explores the impact of different deployment regions, energy sources, and optimization metrics.

\textbf{Execution Setup:}
Bayesian optimization is performed on a single node (8×V100 GPUs, 80 CPUs) over 100 trials, taking 5-20 hours depending on the model. Post-pruning training of CarbonCLIP uses 224 GPUs on the MetaCLIP-2.5B dataset~\cite{metaclip}, with a batch size of 128, learning rate of $5 \times 10^{-4}$, and 2 distillation epochs.
We quantify the Carbon footprint of running~\ourframework in Appendix~\ref{app:carbonFramework}.

\subsection{Joint Model and Hardware Architecture Search Using Different Metrics}
\label{sec:joint}

\begin{figure}
\centering
\includegraphics[trim={0.5cm 0.5cm 0cm 0cm},clip,width=\columnwidth]{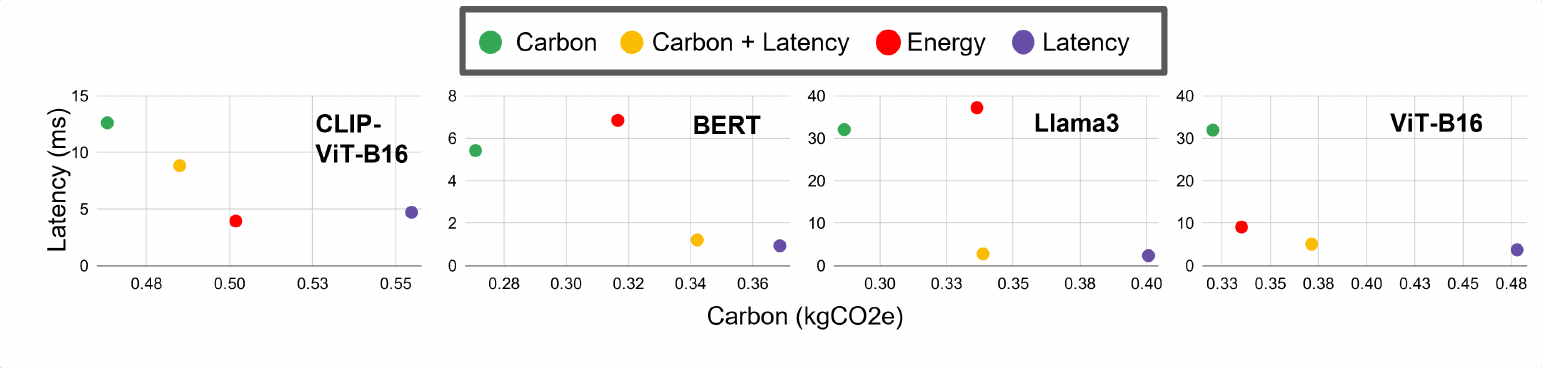}
\vspace{-2ex}
\caption{ISO-Accuracy plot showing the latency–carbon trade-off across optimization strategies, with accuracy matched within ±1\%.}
\label{fig:isoaccuracy}
\vspace{-3ex}
\end{figure}

In this section, we use \ourframework to perform joint model–hardware architecture search, with carbon footprint as a central design metric alongside traditional objectives like accuracy, latency, and energy.
We evaluate different optimization modes under a 20 TOPS compute budget, representative of modern edge accelerators~\cite{hailo,jetson-orin}. When latency is
not an optimization target, a maximum latency constraint of 50ms is enforced~\cite{latency} to ensure realistic specifications. Each experiment is repeated three times, and we compute the Hypervolume (HV) indicator to assess consistency. HV measures the portion of the objective space dominated by the Pareto front relative to a reference point. Across runs, its standard deviation is below 0.03, with an average coefficient of variation under 3.5\%, indicating statistically consistent results. Appendix~\ref{app:consistency} provides a detailed breakdown of the statistics.

\textbf{Takeaway 1:}  \textit{Carbon optimization yields the lowest footprint but at the cost of latency; energy optimization strikes a better balance.} 

Figure~\ref{fig:isoaccuracy} shows the latency-carbon trade-off at iso-accuracy points (based on proxy accuracy before fine-tuning). Full configurations across additional accuracy levels for each model are detailed in Appendix~\ref{app:isoaccuracy}.
Comparing optimization modes, we find that carbon optimization significantly reduces carbon footprint by an average of 30\%, but with a 7.7$\times$ increase in latency compared to latency-optimized baselines. Energy-optimized configurations reduce carbon by 24\% while limiting latency overhead to 4$\times$.
Joint optimization for Carbon and Latency achieves a 18\% carbon reduction with minimal latency increase, effectively balancing sustainability and performance.

\textbf{Takeaway 2:} \textit{Energy optimization indirectly reduces latency, but not as effective as direct latency optimization.}

Minimizing energy tends to reduce both power and computation time, often lowering latency. In contrast, carbon-focused optimization emphasizes minimizing hardware area to reduce embodied carbon, often resulting in slower, smaller designs.
Because total energy consumption is proportional to latency (i.e., energy = power × delay), latency cannot be extended arbitrarily when using smaller-area hardware. Excessively long latencies not only fail to meet practical performance targets but also begin to significantly increase operational carbon costs.
Still, energy optimization does not always guarantee low latency. For example, Bert and Llama3 configurations in Figure~\ref{fig:isoaccuracy} show higher latency despite energy-focused design. Direct latency optimization, seen in Latency-only and Carbon+Latency modes, consistently delivers the lowest-latency results across all models.
These trade-offs highlight the need to carefully balance carbon constraints and real-world performance demands.

\textbf{Takeaway 3:}
\textit{Latency-optimized designs favor large, high-throughput hardware; carbon-optimized designs use compact, low-power accelerators.}

Latency-focused designs often use up to 3$\times$ larger area, enabled by more compute units and larger memory hierarchies. These systems frequently select larger models that fully utilize hardware to meet latency targets.  In contrast, energy-optimized designs typically pair smaller models with larger accelerators to minimize delay and energy per inference.
Carbon-optimized designs prioritize smaller hardware to reduce embodied carbon, at the cost of performance.
For example, on BERT-Base (Figure~\ref{fig:isoaccuracy}), the carbon-optimized system uses a single-core accelerator with 512 Processing Elements (PEs), 64KB local memory, and 2MB global memory.
The latency-optimized design uses two cores, each with 4K PEs, 128KB local memory, and 4MB global memory—cutting latency by over 4$\times$ but increasing carbon footprint by 26\%.

\textbf{Takeaway 4:} \textit{Model sensitivity to pruning varies by model architecture: CLIP and ViT are especially sensitive to hidden dimension pruning.}

Each model architecture responds differently to pruning. CLIP and ViT models are particularly sensitive to reductions in the hidden dimension, leading \ourframework to typically follow this pruning order for these models: FFN dimension $\rightarrow$ attention heads $\rightarrow$ layers $\rightarrow$ embedding dimension. This sensitivity arises because ViTs split images into relatively few patch tokens, requiring each token to carry rich semantic and spatial information. The hidden dimension governs the expressiveness of these token embeddings, and pruning it severely limits ViT's ability to model visual content—especially since ViTs lack the hierarchical features of CNNs and rely on global attention.

CLIP models are especially sensitive due to their multi-modal nature, where precise alignment between visual and textual embeddings is crucial. Pruning hidden or embedding dimensions can significantly degrade performance (Appendix~\ref{app:pruning}), as excessive pruning of the vision encoder disrupts alignment with the text encoder, harming model accuracy.
Additionally, the text encoder is often pruned more aggressively than the vision encoder, which is essential for processing complex visual inputs. 
In contrast, language models such as Bert and Llama3 tend to be more robust to pruning in the hidden dimension.
These observations underscore the challenges of pruning multi-modal models like CLIP and motivate our focus on their optimization throughout the paper.


\subsection{Hardware Optimization for Fixed CLIP Models}
\begin{scriptsize}
\begin{table*}[t]
\centering
\caption{Hardware architecture search for fixed baselines CLIP model architectures.}
\renewcommand{\arraystretch}{1.1} %
\resizebox{1\textwidth}{!}
{\begin{tabular}{lccccccccccc}
\hline
\multirow{3}{*}{Model Architecture} & \multirow{3}{*}{\begin{tabular}[c]{@{}c@{}}Total \\ Params \\ (M)\end{tabular}} & \multicolumn{5}{c}{Minimum Carbon} & \multicolumn{5}{c}{Minimum Latency} \\ \cline{3-12} 
 &  & \multicolumn{1}{c}{\multirow{2}{*}{\begin{tabular}[c]{@{}c@{}}Carbon\\ (kgCO2e)\end{tabular}}} & \multicolumn{1}{c}{\multirow{2}{*}{\begin{tabular}[c]{@{}c@{}}Latency\\ (ms)\end{tabular}}} & \multicolumn{3}{c}{Hardware Architecture} & \multicolumn{1}{c}{\multirow{2}{*}{\begin{tabular}[c]{@{}c@{}}Carbon\\ (kgCO2e)\end{tabular}}} & \multicolumn{1}{c}{\multirow{2}{*}{\begin{tabular}[c]{@{}c@{}}Latency\\ (ms)\end{tabular}}} & \multicolumn{3}{c}{Hardware Architecture} \\ \cline{5-7} \cline{10-12} 
 &  & \multicolumn{1}{c}{} & \multicolumn{1}{c}{} & \multicolumn{1}{c}{\# Cores} & \multicolumn{1}{c}{\begin{tabular}[c]{@{}c@{}}Core\\ Dimension\end{tabular}} & \begin{tabular}[c]{@{}c@{}}Memory Config \\ \{Local, Global\}\end{tabular} & \multicolumn{1}{c}{} & \multicolumn{1}{c}{} & \multicolumn{1}{c}{\# Cores} & \multicolumn{1}{c}{\begin{tabular}[c]{@{}c@{}}Core\\ Dimension\end{tabular}} & \begin{tabular}[c]{@{}c@{}}Memory Config \\ \{Local, Global\}\end{tabular} \\ \hline
CLIP-B/16 & 149 & \multicolumn{1}{c}{0.54} & \multicolumn{1}{c}{18.5} & \multicolumn{1}{c}{1} & \multicolumn{1}{c}{(256,8)} & 64 KB, 2MB & \multicolumn{1}{c}{0.69} & \multicolumn{1}{c}{5.4} & \multicolumn{1}{c}{2} & \multicolumn{1}{c}{(256,16)} & 256KB, 4MB \\
CLIP-L/14 & 427 & \multicolumn{1}{c}{1.43} & \multicolumn{1}{c}{68.7} & \multicolumn{1}{c}{2} & \multicolumn{1}{c}{(128,16)} & 128 KB, 2MB & \multicolumn{1}{c}{1.76} & \multicolumn{1}{c}{66.4} & \multicolumn{1}{c}{4} & \multicolumn{1}{c}{(64,64)} & 256KB, 4MB \\
CLIP-H/14 & 986 & \multicolumn{1}{c}{1.92} & \multicolumn{1}{c}{71.0} & \multicolumn{1}{c}{1} & \multicolumn{1}{c}{(128,32)} & 128KB, 4MB & \multicolumn{1}{c}{2.60} & \multicolumn{1}{c}{70.2} & \multicolumn{1}{c}{4} & \multicolumn{1}{c}{(256,4)} & 512KB, 4MB \\ \hline
TinyCLIP-8M/16 & 41 & \multicolumn{1}{c}{0.34} & \multicolumn{1}{c}{3.0} & \multicolumn{1}{c}{2} & \multicolumn{1}{c}{(256,4)} & 64KB, 2MB & \multicolumn{1}{c}{0.56} & \multicolumn{1}{c}{1.3} & \multicolumn{1}{c}{1} & \multicolumn{1}{c}{(256,64)} & 256KB, 8MB \\
TinyCLIP-39M/16 & 83 & \multicolumn{1}{c}{0.44} & \multicolumn{1}{c}{9.4} & \multicolumn{1}{c}{1} & \multicolumn{1}{c}{(256,8)} & 64KB, 2MB & \multicolumn{1}{c}{0.59} & \multicolumn{1}{c}{2.2} & \multicolumn{1}{c}{4} & \multicolumn{1}{c}{(256,16)} & 128KB, 4MB \\
TinyCLIP-40M/32 & 84 & \multicolumn{1}{c}{0.37} & \multicolumn{1}{c}{8.6} & \multicolumn{1}{c}{1} & \multicolumn{1}{c}{(32,32)} & 64KB, 2MB & \multicolumn{1}{c}{0.46} & \multicolumn{1}{c}{1.1} & \multicolumn{1}{c}{4} & \multicolumn{1}{c}{(128,32)} & 64KB, 2MB \\
TinyCLIP-61M/32 & 115 & \multicolumn{1}{c}{0.39} & \multicolumn{1}{c}{9.7} & \multicolumn{1}{c}{1} & \multicolumn{1}{c}{(128,8)} & 64KB, 2MB & \multicolumn{1}{c}{0.49} & \multicolumn{1}{c}{1.4} & \multicolumn{1}{c}{4} & \multicolumn{1}{c}{(128,32)} & 64KB, 2MB \\ \hline
\end{tabular}}
\label{tab:baselines}
\end{table*}
\end{scriptsize}
\label{sec:clip_hw_only}
While the previous section highlighted \ourframework\unskip’ general utility, we now focus on CLIP—a widely used multi-modal model combining text and vision encoders. CLIP poses additional challenges due to its heterogeneous computation and the complex interplay between modality-specific pruning and system performance.
We first evaluate the carbon footprint of state-of-the-art CLIP variants. To ensure fair comparison, we use \ourframework\unskip’ hardware search to optimize inference latency and carbon footprint under a fixed model and a 20 TOPS compute budget. This yields a Pareto frontier illustrating optimal latency–carbon trade-offs. Table~\ref{tab:baselines} summarizes configurations with minimum carbon and minimum latency for each model.
As with joint optimization, carbon-focused hardware design results in smaller accelerators with fewer compute and memory resources. While this reduces area and emissions, it typically increases inference latency.

\textbf{Takeaway 5:} \textit{Hardware must be tailored to the model-parameter count, patch size, and modality-specific traits affect optimal configurations.}

Results from the fixed-model hardware search underscore the importance of hardware–model co-design: optimal performance and carbon efficiency require aligning hardware with model architecture. ViT patch size notably influences accelerator dimensions. For instance, TinyCLIP-61M/32—despite having more parameters—outperforms TinyCLIP-39M/16 in both latency and carbon footprint when paired with hardware better suited to its larger patch size. ViT-B/32 produces a sequence length of 50, compared to 197 for ViT-B/16, enabling more efficient execution on smaller accelerators. Embedding dimensions and patch sizes shape execution patterns, making some models inherently better matched to specific hardware.
Memory configuration also scales with model size: smaller models perform well with 64KB local and 2MB global buffers, while larger models like CLIP-L/14 and CLIP-H/14 require at least 128KB local buffers and larger global memory.

\subsection{Accuracy Evaluation}
\label{sec:accuracy}

We compare CarbonCLIP models to TinyCLIP and the CLIP-ViT-B/16 baseline (pretrained on DataComp-1B), using each baseline’s most carbon-efficient configuration. Table~\ref{tab:carbonclip} shows the CarbonCLIP family, selected from the Pareto frontiers in Appendix~\ref{app:pareto}. For each CarbonCLIP model, We perform post-pruning training, and evaluate their performance on 41 zero-shot tasks from the CLIPBenchmark~\cite{clip_benchmark}.
The table summarizes model and hardware configurations for CarbonCLIP-XL to CarbonCLIP-XS (largest to
smallest models), along with their carbon footprint, latency, and average accuracy. Full per-dataset results are in Appendix~\ref{app:carbonclip}. We also extend the CarbonCLIP family to the CLIP-B/32 architecture for comparison with TinyCLIP-B/32 baselines (Appendix~\ref{app:b32}).

\textbf{CarbonCLIP models outperform baselines across a range of carbon budgets.} Notably, CarbonCLIP-XL achieves baseline-level accuracy with an 10\% reduction in carbon footprint. CarbonCLIP-XS achieves an 8\% increase in accuracy with a 3\% reduction in carbon footprint compared to TinyCLIP-8M/16. CarbonCLIP-L, CarbonCLIP-M, and CarbonCLIP-S all achieve significant reductions in carbon footprint compared to TinyCLIP-39M/16, with CarbonCLIP-L achieving a 4\% increase in accuracy and a 4.5\% reduction in carbon footprint, CarbonCLIP-M achieving an 11\% reduction in carbon footprint with a 3\% decrease in accuracy, and CarbonCLIP-S achieving a 17\% reduction in carbon footprint without any regression in accuracy.

In terms of hardware configuration, CarbonCLIP models select accelerators with cores of $PE_x$ dimension 256 to align with the underlying operator dimensions of the CLIP ViT-B/16 architecture, with sequence length of 197 for the vision encoder. Smaller models select a total of 1024 PE units per core, whereas larger models select twice as many PEs to keep the latency of the task low. Due to the reduced size of the CarbonCLIP models, a 64KB local memory and 2MB global memory are sufficient to keep the core utilized.

\begin{scriptsize}
\begin{table*}[t]
\caption{Hardware and model architecture properties of each variant of the CarbonCLIP family. Hardware configurations are specified as: $\{\mathrm{TC}, \mathrm{PE}_x, \mathrm{PE}_y, \mathrm{L2}, \mathrm{L2}_{bw}, \mathrm{GLB} \}$. $\mathrm{PE}$ denotes Processing Element. Text and Vision encoders specified as: \{Num Layers, FFN Dim, Hidden Dim, Num Heads\}.}
\renewcommand{\arraystretch}{1.1} %
\resizebox{1\textwidth}{!}
{\begin{tabular}{lccccccc}
\hline
\multirow{2}{*}{Name} & \multirow{2}{*}{\begin{tabular}[c]{@{}c@{}}Carbon\\ (kgCO2e)\end{tabular}} & \multirow{2}{*}{\begin{tabular}[c]{@{}c@{}}Latency\\ (ms)\end{tabular}} & \multirow{2}{*}{\begin{tabular}[c]{@{}c@{}}Hardware\\ Configuration\end{tabular}} & \multicolumn{3}{c}{\begin{tabular}[c]{@{}c@{}}Model Configuration\end{tabular}} & \multirow{2}{*}{\begin{tabular}[c]{@{}c@{}}Avg. Accuracy \\ over 41 datasets\end{tabular}} \\ \cline{5-7}
 &  &  &  & \multicolumn{1}{c}{\begin{tabular}[c]{@{}c@{}}Text Encoder\\ Configuration\end{tabular}} & \multicolumn{1}{c}{\begin{tabular}[c]{@{}c@{}}Vision Encoder\\ Configuration\end{tabular}} & Params (M) &  \\ \hline
CLIP-B/16 - DataComp & 0.54 & 18.5 & \{1, 256, 8, 64, 128, 2\} & \multicolumn{1}{c}{\{12, 2048, 512, 8\}} & \multicolumn{1}{c}{\{12, 3072, 768, 12\}} & 149 & 53.2 \\ \hline
TinyCLIP-8M/16 & 0.34 & 3.0 & \{2, 256, 4, 64, 32, 2 \} & \multicolumn{1}{c}{\{3, 1024, 256, 4\}} & \multicolumn{1}{c}{\{10, 1024, 256, 4\}} & 41 & 30.7 \\
TinyCLIP-39M/16 & 0.44 & 9.4 & \{1, 256, 8, 64, 128, 2\} & \multicolumn{1}{c}{\{6, 2048, 512, 8\}} & \multicolumn{1}{c}{\{12, 2048, 512, 8\}} & 83 & 45.0 \\ \hline
\textbf{CarbonCLIP-XS} & 0.32 & 7.1 & \{1, 256, 4, 64, 32, 2\} & \multicolumn{1}{c}{\{6, 1024, 284, 4\}} & \multicolumn{1}{c}{\{6, 1536, 576, 6\}} & 41 & 38.7 \\
\textbf{CarbonCLIP-S} & 0.36 & 12.0 & \{1, 256, 4, 64, 64, 2\} & \multicolumn{1}{c}{\{6, 1024, 512, 6\}} & \multicolumn{1}{c}{\{8, 1920, 672, 6\}} & 63 & 45.0 \\
\textbf{CarbonCLIP-M} & 0.39 & 19.7 & \{1, 256, 4, 64, 128,2 \} & \multicolumn{1}{c}{\{8, 1536, 512, 6\}} & \multicolumn{1}{c}{\{9, 2304, 672, 6\}} & 79 & 47.9 \\
\textbf{CarbonCLIP-L} & 0.42 & 13.7 & \{1, 256, 8, 64, 128, 2\} & \multicolumn{1}{c}{\{6, 1280, 384, 5\}} & \multicolumn{1}{c}{\{10, 2688, 768, 7\}} & 83 & 48.7 \\
\textbf{CarbonCLIP-XL} & 0.49 & 19.1 & \{1, 256, 8, 64, 128, 2\} & \multicolumn{1}{c}{\{12, 2048, 512, 4\}} & \multicolumn{1}{c}{\{12, 3072, 768, 5\}} & 123 & 52.0 \\ \hline
\end{tabular}}
\label{tab:carbonclip}
\end{table*}
\end{scriptsize}

\subsection{Evaluating for Different Compute Constraints}
\label{sec:compute_constraints}

\begin{figure*}[t]
    \centering
    \includegraphics[width=0.9\textwidth]{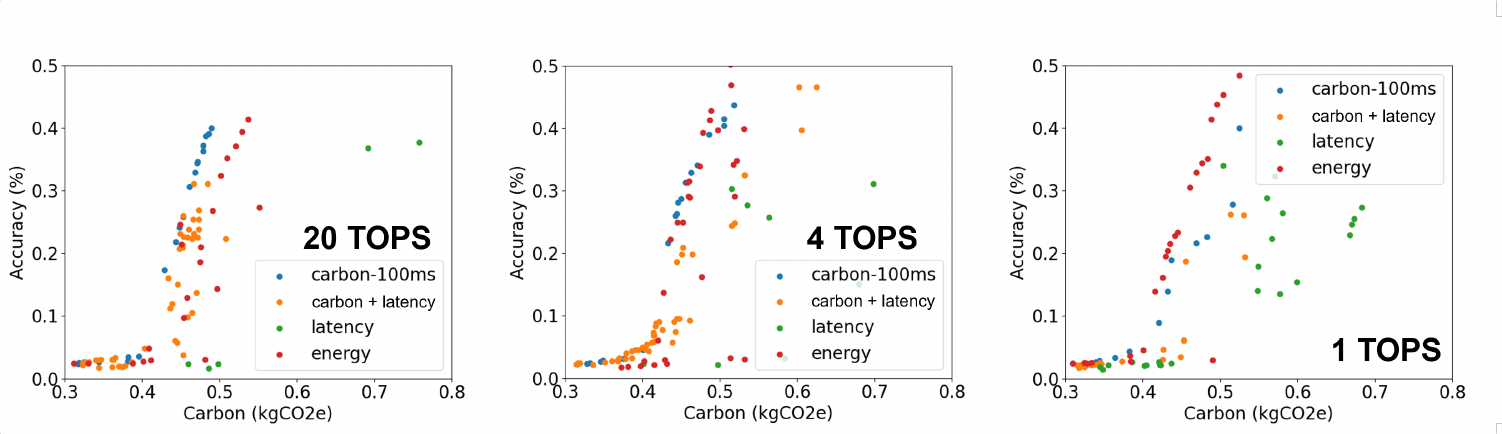}
    \vspace{-0.2cm}
    \caption{Pareto frontiers under varying compute constraints.}
    \label{fig:compute}
    \vspace{-2ex}
\end{figure*}

We evaluate CarbonCLIP models under three peak compute constraints inspired by real-world edge devices: 20 TOPS~\cite{hailo,jetson-orin}, 4 TOPS~\cite{edgetpu}, and 1 TOPS~\cite{3d, myriad}. Figure~\ref{fig:compute} shows the resulting Pareto frontiers.

\textbf{Takeaway 6:}  \textit{As hardware shrinks due to tighter compute constraints, optimizing for energy (operational carbon) increasingly minimizes both total carbon and latency.} 

Smaller hardware designs reduce embodied carbon, making operational energy the dominant factor in the total carbon footprint. As a result, energy-optimized designs often achieve outcomes similar to those optimized for total carbon under tight compute budgets. Appendix~\ref{app:breakdown} explores the embodied vs. operational carbon trade-offs in more detail. Note that different grid intensities and
expected hardware lifetime can also affect the ratio of embodied and operational carbon, affecting the optimization results, and the quality
of each metric immensely, we show this in Appendix~\ref{app:region}.
However, smaller accelerators have limited compute and memory, which increases inference delay—and thus total energy use—despite lower power consumption. This leads to higher total carbon emissions in some cases, underscoring the importance of hardware–model co-design to balance carbon efficiency and performance. We further extended the study to evaluate the impact of latency constraints on model and hardware configurations, as well as their carbon footprints in Appendix~\ref{app:latency}. 



\section{Impact Discussion and Limitations}
\label{sec:limitations}



This work demonstrates a practical path toward sustainable AI by jointly optimizing model and hardware design for carbon reduction. Our framework is particularly effective for low-latency edge inference (e.g., chatbots, AR/VR) and highlights the potential for cross-domain collaboration across ML, hardware, and sustainability.  

\ourframework currently targets Transformer models and domain-specific edge accelerators, with limited GPU support and no evaluation on CNNs. Extending to other model families (e.g., CNNs, Mamba) requires profiling the latency and energy of their unique operators and adapting pruning strategies beyond attention heads. Expanding to GPUs and data center scenarios is feasible, but the larger design space would demand significantly more search. Nonetheless, our GPU-based results (Appendices~\ref{app:model_gpu} and \ref{app:gpu_validation}) suggest scalability to training workloads and datacenter-level accelerators, where support for parallelization and distributed training could further improve carbon efficiency.  

Beyond inference and training emissions, broader sustainability impacts, such as electronic waste, water usage, and rare mineral consumption, remain underexplored. While our optimizations can implicitly reduce chip area and resource use, standardized metrics and fine-grained data are currently lacking, limiting multi-objective optimization over broader environmental impacts. Moreover, custom accelerators raise concerns about hardware heterogeneity and e-waste, which may be mitigated by reusing existing hardware or co-optimizing with commercial accelerators.  

Proxy-based accuracy estimation, such as the methods employed in \ourframework, enables scalable exploration but may deviate from real outcomes due to overfitting or underestimating pruning effects. Developing more robust accuracy predictors is an important direction for improving reliability.

\section{Conclusion}

We introduced \ourframework, a framework for co-optimizing model architectures and domain-specific accelerators to minimize carbon footprint. By jointly considering operational and embodied carbon, \ourframework ~supports environmentally conscious design, particularly in edge computing. We demonstrated its effectiveness across various Transformer-based models—including multi-modal models—showcasing substantial potential in carbon reductions without sacrificing performance. This work fills a key gap in sustainable AI deployment and provides a foundation for future research in carbon-aware machine learning.

\section{Acknowledgements}
We would like to thank Bernie Beckerman, David Eriksson, and Max Balandat for their expertise and assistance in building the optimization platform with Ax, Igor Fedorov for sharing insights on pruning models, Daniel Li and Hu Xu for providing valuable guidance on training CLIP models and working with the MetaCLIP dataset. We also would like thank Kim Hazelwood, Kristen Lauter, and Edith Beigne for supporting this work.

\newpage
\bibliography{example_paper}
\bibliographystyle{alpha}

\newpage
\newpage
\appendix
\onecolumn
\section*{Appendix}
\section{Ablation Study: Pruning and Finetuning CLIP models}
\label{app:pruning}

\begin{figure*}[h]
\centering
\includegraphics[width=1\textwidth]{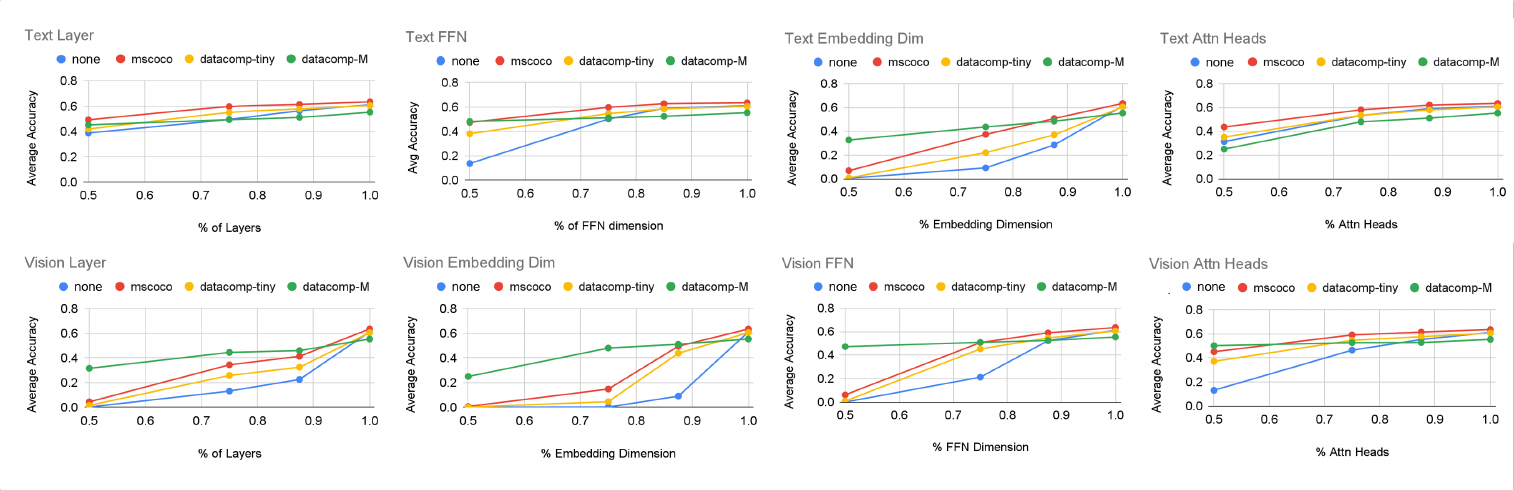}
\caption{Pruning and fine-tuning results for each dimension of the Text and Vision encoders.}
\label{fig:pruning_all}
\vspace{-3ex}
\end{figure*}

In this ablation study, we studied the effect of pruning each dimension and the effect of fine-tuning on various datasets to find a good proxy for approximating the accuracy after training the pruned models on larger datasets fir CLIP-based models. All models were evaluated against MS COCO. The results are shown in Figure~\ref{fig:pruning_all} and our key observations are as follows:

We make some key observations in terms of the importance of each dimension to the overall accuracy of the model. First, accuracy drops significantly after pruning to 50\% of any dimension, even after fine-tuning. Therefore, we confine the search space to a minimum of half of each dimension. Second, the vision model has a more significant impact on accuracy compared to the text model. Finally, pruning the embedding dimension has the most significant impact on accuracy among all pruned parameters, followed by number of layers, then FFN dimensions and number of attention heads.

\textbf{Fine-tuning on MS COCO:} Even with just a single epoch, fine-tuning on MS COCO significantly improves the accuracy of pruned models, making it a good and fast proxy for evaluating their overall potential.
\textbf{Fine-tuning on Datacomp-Tiny:} Training on a small subset of a pre-training dataset (Datacomp-Tiny, a 400k subset of Datacomp-Medium unfiltered) also improves accuracy, albeit with lower overall accuracy compared to MS COCO.
\textbf{Fine-tuning on Datacomp-Medium:} Using a general pre-training dataset (Datacomp-Medium unfiltered) reduces the variance in accuracy between models of different sizes, showing that smaller models can achieve comparable accuracy when trained with enough data. This incentivizes our post-pruning training with a large and high quality MetaCLIP 2.5B Dataset.

In general, larger models will attain higher accuracy compared to models with fewer parameters when trained for the same number of steps. However, models with fewer parameters may still achieve comparable accuracy as larger models given more training steps~\cite{scaling}. Therefore, we fine-tune each model with the same computation FLOPS, allowing smaller models to train for more FLOPS and recover their accuracy from more extensive pruning.

\section{Estimation Tool-chain Integration}
\label{app:estimation_tools}

\begin{figure*}
\centering
\includegraphics[width=0.9\textwidth]{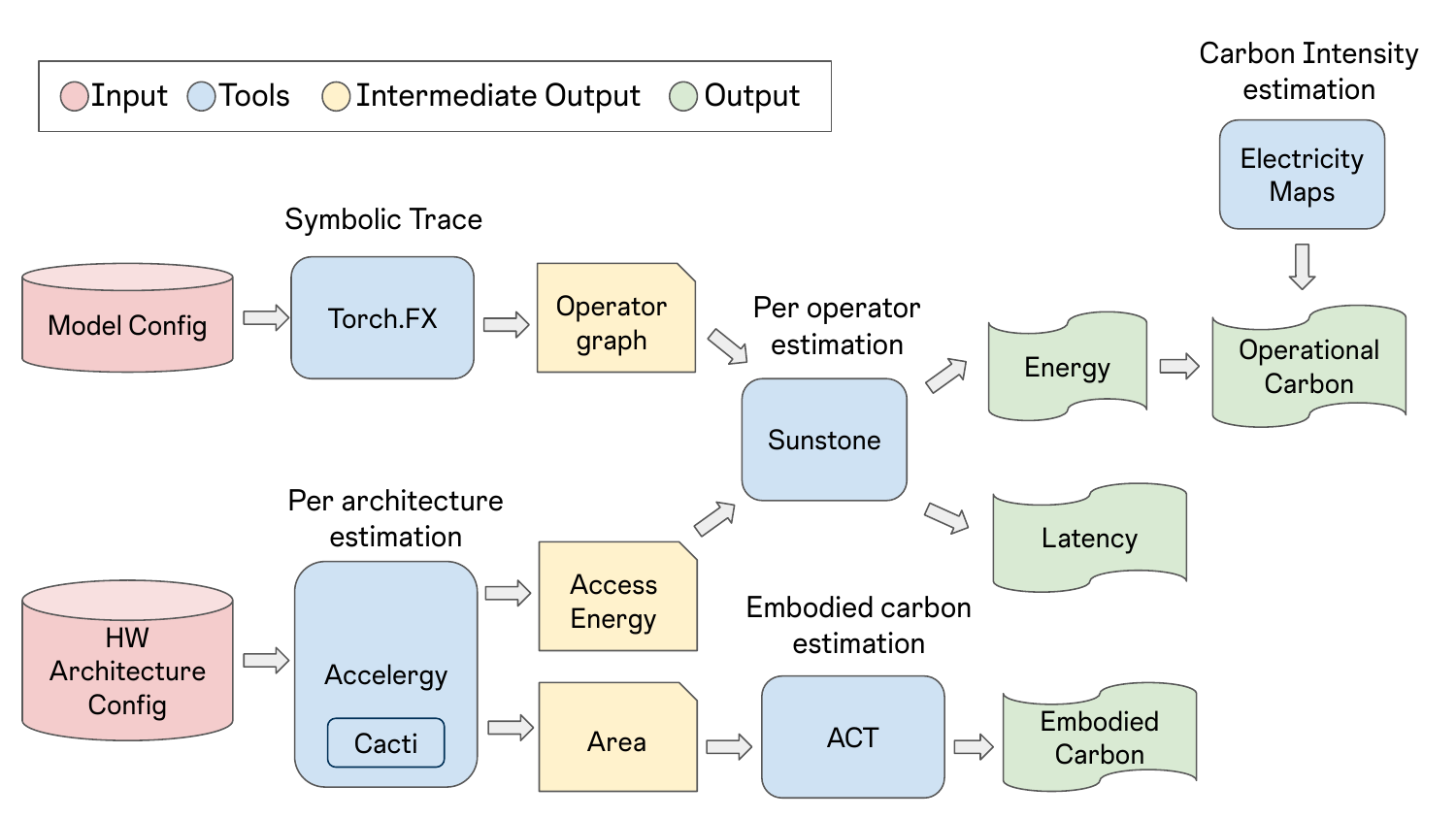}
\caption{Integration of tools to estimate the total carbon footprint of the given hardware and model configuration.}
\label{fig:estimation_tools}
\vspace{-3ex}
\end{figure*}

In this section we present in detail the integration of the tools used in the hardware estimator as illustrated in Figure~\ref{fig:estimation_tools}.
We use Accelergy~\cite{accelergy} to estimate the area and access energy of each component, and Sunstone~\cite{sunstone} to estimate the per-operator latency and energy. For carbon estimations, we use ACT~\cite{act} to estimate the embodied carbon of the hardware architecture based on the area of the accelerator. Given the energy estimates from Sunstone, Electricity Maps~\cite{electricitymaps} is used to retrieve the carbon intensity of the electricity of a given grid location and calculate the operational carbon of executing a single inference. We then scale the operational carbon to the total lifetime of the hardware architecture.

\section{Correlation Between Proxy and Final Model Accuracy}
\label{app:proxy}
\begin{figure}
\centering
\includegraphics[width=0.5\columnwidth]{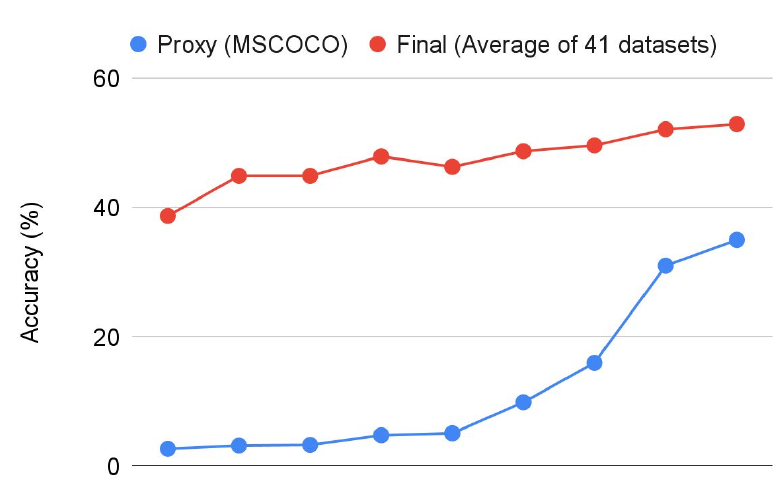}

\caption{Correlation between the proxy accuracy, fine-tuned and evaluated on MS COCO and the final model accuracy evaluated across 41 datasets for pruned CLIP models.}
\label{fig:proxy}
\vspace{-2ex}
\end{figure}

To efficiently approximate accuracy while preserving ranking consistency, we fine-tune each pruned model to a common dataset used to evaluate each model type.  As shown in Figure~\ref{fig:proxy}, fine-tuning maintains a high Spearman’s rank correlation coefficient (0.98) with the final post-pruning training accuracy, ensuring reliable accuracy ranking. We report the mean top-1 recall accuracy on MSCOCO as the accuracy proxy for each pruned CLIP model.
We incorporate this into the optimization loop to dramatically reduce the cost of evaluating candidate model architectures.

\section{ISO Accuracy}
\label{app:isoaccuracy}
In this sections we present full details on different design points found by each optimization metric for a given accuracy point in Tables~\ref{tab:isoaccuracy-clip},~\ref{tab:isoaccuracy-bert},~\ref{tab:isoaccuracy-llama}, and~\ref{tab:isoaccuracy-vit}, for CLIP-ViT-B16, BertBase, Llama3-8b, ViT-B16, respectively. The key observations are discussed in Section~\ref{sec:joint}. For each iso-accuracy comparison, accuracy values are reported with a tolerance of +/- 1\% across optimization strategies.


\begin{scriptsize}

\begin{table}[]
\centering
\caption{The hardware and model architecture configuration found by each optimization metric at each accuracy point for CLIP-ViT-B16. Hardware configurations are specified in the format of: $\{TC, pe_x, pe_y, L2, L2_{bw}, glb \}$. Text and Vision encoders are specified in the format of \{Num Layers, FFN Dim, Hidden Dim, Num Heads\}}
\label{tab:isoaccuracy-clip}
\resizebox{1\columnwidth}{!}
{\begin{tabular}{cccccccc}
\hline
\multirow{2}{*}{\begin{tabular}[c]{@{}c@{}}Accuracy\\ (+/- 1\%)\end{tabular}} & \multirow{2}{*}{\begin{tabular}[c]{@{}c@{}}Optimization \\ Metric\end{tabular}} & \multirow{2}{*}{\begin{tabular}[c]{@{}c@{}}Carbon\\ (kgCO2e)\end{tabular}} & \multirow{2}{*}{\begin{tabular}[c]{@{}c@{}}Latency\\ (ms)\end{tabular}} & \multirow{2}{*}{\begin{tabular}[c]{@{}c@{}}Hardware\\ Configuration\end{tabular}} & \multicolumn{3}{c}{\begin{tabular}[c]{@{}c@{}}Model\\ Configuration\end{tabular}} \\ \cline{6-8} 
 &  &  &  &  & \multicolumn{1}{c}{\begin{tabular}[c]{@{}c@{}}Text Encoder\\ Configuration\end{tabular}} & \multicolumn{1}{c}{\begin{tabular}[c]{@{}c@{}}Vision Encoder\\ Configuration\end{tabular}} & Params (M) \\ \hline
\multirow{4}{*}{31\%} & Carbon & 0.46 & 12.6 & \{1, 256, 8, 64, 64, 2\} & \multicolumn{1}{c}{\{9, 1536, 512, 6\}} & \multicolumn{1}{c}{\{12, 576, 768, 8\}} & 104 \\
 & Energy & 0.50 & 3.9 & \{2, 256, 16, 128, 32, 2\} & \multicolumn{1}{c}{\{7, 1536, 384, 8\}} & \multicolumn{1}{c}{\{12, 576, 768, 8\}} & 101 \\
 & Latency & 0.55 & 4.7 & \{2, 256, 16, 128, 32, 2\} & \multicolumn{1}{c}{\{10, 1792, 384, 7\}} & \multicolumn{1}{c}{\{12, 672, 768, 11\}} & 111 \\
 & Carbon + Latency & 0.48 & 8.8 & \{2, 256, 8, 64, 64, 2\} & \multicolumn{1}{c}{\{9, 1280, 512, 6\}} & \multicolumn{1}{c}{\{11, 672, 768, 9\}} & 105 \\ \hline
\multirow{4}{*}{19.5\%} & Carbon & 0.44 & 10.9 & \{1,256, 8, 64, 64 2\} & \multicolumn{1}{c}{\{9, 1536, 512, 6\}} & \multicolumn{1}{c}{\{11, 672, 768,6\}} & 95 \\
 & Energy & 0.48 & 3.5 & \{2, 256, 16, 128, 32, 2\} & \multicolumn{1}{c}{\{6, 1536, 512, 8\}} & \multicolumn{1}{c}{\{12, 480, 768, 7\}} & 90 \\
 & Latency & 0.55 & 8.2 & \{4, 256, 4, 128, 64, 2\} & \multicolumn{1}{c}{\{11, 1536, 384, 5\}} & \multicolumn{1}{c}{\{11, 576, 768, 12\}} & 99 \\
 & Carbon + Latency & 0.45 & 7.3 & \{2, 256, 4, 64, 128, 2\} & \multicolumn{1}{c}{\{9, 1024, 512, 5\}} & \multicolumn{1}{c}{\{11, 576, 768, 8\}} & 94 \\ \hline
\multirow{4}{*}{13\%} & Carbon & 0.43 & 22.1 & \{1, 256, 4, 64, 64, 2\} & \multicolumn{1}{c}{\{7, 1536, 6384 5\}} & \multicolumn{1}{c}{\{11, 576, 768, 8\}} & 84 \\
 & Energy & 0.49 & 7.3 & \{2, 256, 8, 64, 64, 2\} & \multicolumn{1}{c}{\{8, 1792, 448, 5\}} & \multicolumn{1}{c}{\{10, 576, 768, 7\}} & 92 \\
 & Latency & 0.54 & 15.9 & \{4, 256, 2, 128, 64, 2\} & \multicolumn{1}{c}{\{12, 1792, 320, 5\}} & \multicolumn{1}{c}{\{11, 576, 768, 12\}} & 96 \\
 & Carbon + Latency & 0.47 & 7.3 & \{1, 256, 16, 128, 64, 2\} & \multicolumn{1}{c}{\{9, 2048, 384, 4\}} & \multicolumn{1}{c}{\{11, 3072, 768, 6\}} & 98 \\ \hline
\multirow{4}{*}{2.5\%} & Carbon & 0.32 & 4.6 & \{1, 256, 8, 64, 64, 2\} & \multicolumn{1}{c}{\{6, 1024, 256, 4\}} & \multicolumn{1}{c}{\{6, 1536, 384, 6\}} & 27 \\
 & Energy & 0.33 & 1.8 & \{1, 256, 16, 128, 64, 2\} & \multicolumn{1}{c}{\{6, 1024, 256, 6\}} & \multicolumn{1}{c}{\{6, 1536, 384, 6\}} & 28 \\
 & Latency & 0.46 & 1.3 & \{4, 256, 16, 128, 128, 2\} & \multicolumn{1}{c}{\{6, 1024, 384, 8\}} & \multicolumn{1}{c}{\{9, 1536, 384, 4\}} & 43 \\
 & Carbon + Latency & 0.31 & 5.1 & \{1, 256, 4, 64, 64, 2\} & \multicolumn{1}{c}{\{6, 1024, 256, 4\}} & \multicolumn{1}{c}{\{6, 1536, 384, 6\}} & 27 \\ \hline
\end{tabular}}
\end{table}
\end{scriptsize}

\begin{scriptsize}
\begin{table}[]
\caption{Iso-Accuracy points for Bert-Base model. Original model configuration:  number of layers: 12, embedding dimension: 768, intermediate dimension:3072, number of Attention Heads:12}
\label{tab:isoaccuracy-bert}
\resizebox{1\columnwidth}{!}
{\begin{tabular}{ccccccc}
\hline
\multirow{2}{*}{\begin{tabular}[c]{@{}c@{}}Accuracy\\ (+/- 1\%)\end{tabular}} & \multirow{2}{*}{\begin{tabular}[c]{@{}c@{}}Optimization \\ Metric\end{tabular}} & \multirow{2}{*}{\begin{tabular}[c]{@{}c@{}}Carbon\\ (kgCO2e)\end{tabular}} & \multirow{2}{*}{\begin{tabular}[c]{@{}c@{}}Latency\\ (ms)\end{tabular}} & \multirow{2}{*}{\begin{tabular}[c]{@{}c@{}}Hardware\\ Configuration\end{tabular}} & \multicolumn{2}{c}{Model} \\ \cline{6-7} 
 &  &  &  &  & \multicolumn{1}{c}{Configuration} & Params (M) \\ \hline
\multirow{4}{*}{63\%} & Carbon & 0.26 & 12.7 & \{1, 32, 8, 64, 32, 2\} & \multicolumn{1}{c}{\{10, 192, 48, 6\}} & 1.7 \\
 & Energy & 26 & 2.9 & \{4, 1, 62, 8, 64, 64, 2\} & \multicolumn{1}{c}{\{6, 384, 48, 4\}} & 1.7 \\
 & Latency & 0.41 & 0.9 & \{4, 256, 8, 256, 128, 4\} & \multicolumn{1}{c}{\{9,192,144,6\}} & 5.5 \\
 & Carbon + Latency & 0.31 & 0.9 & \{4, 4, 128, 8, 128, 64, 2\} & \multicolumn{1}{c}{\{8, 576, 48, 4\}} & 1.9 \\ \hline
\multirow{4}{*}{68\%} & Carbon & 0.44 & 10.9 & \{1, 128, 4, 64, 64, 2\} & \multicolumn{1}{c}{\{7, 576, 48, 12\}} & 1.9 \\
 & Energy & 0.48 & 3.5 & \{1, 128, 2, 128, 128\} & \multicolumn{1}{c}{\{8, 1344, 48, 6\}} & 2.5 \\
 & Latency & 0.55 & 8.2 & \{2, 256, 16, 128, 128, 4\} & \multicolumn{1}{c}{\{7, 1920, 48, 6\}} & 2.7 \\
 & Carbon + Latency & 0.45 & 7.3 & \{2, 256, 16, 256, 128, 2\} & \multicolumn{1}{c}{\{6, 3072, 96, 8\}} & 6.4 \\
 \hline
\end{tabular}}
\end{table}
\end{scriptsize}

\begin{table}[]
\caption{Iso-Accuracy points for Llama2-7b model. Original model configuration:  number of layers: 32, intermediate dimension:11008, embedding dimension: 4096, number of Attention Heads:32}
\label{tab:isoaccuracy-llama}
\resizebox{1\columnwidth}{!}
{\begin{tabular}{ccccccc}
\hline
\multirow{2}{*}{\begin{tabular}[c]{@{}c@{}}Accuracy\\ (+/- 1\%)\end{tabular}} & \multirow{2}{*}{\begin{tabular}[c]{@{}c@{}}Optimization \\ Metric\end{tabular}} & \multirow{2}{*}{\begin{tabular}[c]{@{}c@{}}Carbon\\ (kgCO2e)\end{tabular}} & \multirow{2}{*}{\begin{tabular}[c]{@{}c@{}}Latency\\ (ms)\end{tabular}} & \multirow{2}{*}{\begin{tabular}[c]{@{}c@{}}Hardware\\ Configuration\end{tabular}} & \multicolumn{2}{c}{Model} \\ \cline{6-7} 
 &  &  &  &  & \multicolumn{1}{c}{Configuration} & Params (M) \\ \hline
\multirow{4}{*}{64\%} & Carbon & 0.26 & 12.7 & \{1, 64, 8, 64, 32, 2\} & \multicolumn{1}{c}{\{2, 1376, 256, 32\}} & 38.2 \\
 & Energy & 26 & 2.9 & \{1, 64, 8, 64, 128, 4\} & \multicolumn{1}{c}{\{2, 3440, 256, 32\}} & 41.2 \\
 & Latency & 0.41 & 0.9 & \{4, 256, 8, 128, 128, 4\} & \multicolumn{1}{c}{\{2, 1376, 256, 32\}} & 38.2 \\
 & Carbon + Latency & 0.31 & 0.9 & \{4, 128, 16, 64, 128, 2\} & \multicolumn{1}{c}{\{2, 1376, 256, 32\}} & 38.2 \\ \hline
\multirow{4}{*}{66\%} & Carbon & 0.44 & 10.9 & \{1, 64, 8, 64, 64, 2\} & \multicolumn{1}{c}{\{3, 1376, 256, 32\}} & 41.8 \\
 & Energy & 0.48 & 3.5 & \{4, 128, 16, 128, 64, 2\} & \multicolumn{1}{c}{\{3, 1376, 256, 32\}} & 41.8 \\
 & Latency & 0.55 & 8.2 & \{2, 256, 16, 512, 64, 8\} & \multicolumn{1}{c}{\{2, 4128, 256, 32\}} & 42.4 \\
 & Carbon + Latency & 0.45 & 7.3 & \{1, 256, 64, 512, 128, 4\} & \multicolumn{1}{c}{\{3, 1376, 256, 32\}} & 41.8 \\
 \hline
\end{tabular}}
\end{table}

\begin{table}[]
\caption{Iso-Accuracy points for ViT-B-16 model. Original model configuration:  number of layers: 12, intermediate dimension:3072, embedding dimension: 768, number of Attention Heads:12}
\label{tab:isoaccuracy-vit}
\resizebox{1\columnwidth}{!}
{\begin{tabular}{ccccccc}
\hline
\multirow{2}{*}{\begin{tabular}[c]{@{}c@{}}Accuracy\\ (+/- 1\%)\end{tabular}} & \multirow{2}{*}{\begin{tabular}[c]{@{}c@{}}Optimization \\ Metric\end{tabular}} & \multirow{2}{*}{\begin{tabular}[c]{@{}c@{}}Carbon\\ (kgCO2e)\end{tabular}} & \multirow{2}{*}{\begin{tabular}[c]{@{}c@{}}Latency\\ (ms)\end{tabular}} & \multirow{2}{*}{\begin{tabular}[c]{@{}c@{}}Hardware\\ Configuration\end{tabular}} & \multicolumn{2}{c}{Model} \\ \cline{6-7} 
 &  &  &  &  & \multicolumn{1}{c}{Configuration} & Params (M) \\ \hline
\multirow{4}{*}{39\%} & Carbon & 0.26 & 12.7 & \{1, 32, 32, 64, 64, 2\} & \multicolumn{1}{c}{\{12, 1920, 2496, 6\}} & 45.8 \\
 & Energy & 26 & 2.9 & \{1, 128, 32, 64, 128, 2\} & \multicolumn{1}{c}{\{11, 1344, 2496, 12\}} & 31.7 \\
 & Latency & 0.41 & 0.9 & \{2, 256, 32, 256, 128, 4\} & \multicolumn{1}{c}{\{7, 2112, 2688, 6\}} & 31.7 \\
 & Carbon + Latency & 0.31 & 0.9 & \{2, 256, 16, 128, 128, 2\} & \multicolumn{1}{c}{\{8, 1728, 2496, 6\}} & 29.0 \\ \hline
\multirow{4}{*}{40\%} & Carbon & 0.44 & 10.9 & \{2, 128, 8, 128, 64, 2\} & \multicolumn{1}{c}{\{11, 2112, 2688, 8\}} & 49.2 \\
 & Energy & 0.48 & 3.5 & \{1, 256, 16, 64, 64, 2\} & \multicolumn{1}{c}{\{12, 1728, 2496, 12\}} & 39.5 \\
 & Latency & 0.55 & 8.2 & \{2, 256, 16, 64, 64, 8\} & \multicolumn{1}{c}{\{11, 1728, 2496, 8\}} & 43.2 \\
 & Carbon + Latency & 0.45 & 7.3 & \{4, 64, 32, 64, 128, 2\} & \multicolumn{1}{c}{\{12, 1728, 2688, 8\}} & 48.0 \\
 \hline
\end{tabular}}
\end{table}

\section{Pareto Frontier of each Optimization}
\label{app:pareto}
\begin{figure*}[t]
    \centering
    \includegraphics[width=\textwidth]{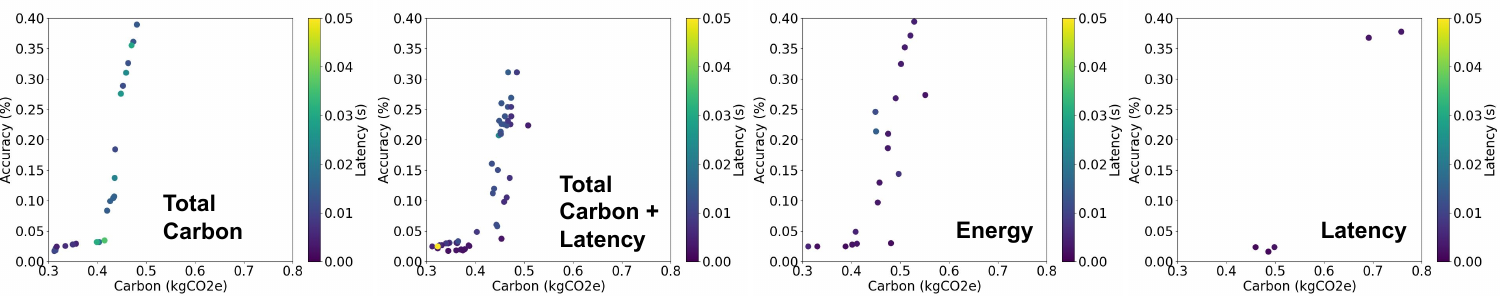}
    \vspace{-4ex}
    \caption{Pareto frontiers for different optimization modes under a 20 TOPS compute constraint. }
    \label{fig:metrics}
\end{figure*}

In this section, we present an example of the Pareto frontier for each optimization metric using the CLIP ViT-B/16 model. Figure~\ref{fig:metrics} shows the Pareto frontiers for latency-only, carbon-only, energy-only, and latency+carbon optimization objectives. Each data point in the figure represents a specific model and hardware architecture configuration, with accuracy shown on the y-axis, carbon footprint on the x-axis, and latency encoded via a color map. To ensure consistency, each experiment is repeated three times, and accuracy is estimated using the MS COCO dataset. When latency is not an explicit optimization objective, a maximum latency constraint of 50 ms is enforced~\cite{latency} to ensure realistic deployment scenarios.

Consistent with the discussion in Section~\ref{sec:joint}, examining the Pareto frontiers reveals that optimizing solely for total carbon yields configurations with the lowest carbon footprint across all accuracy levels. However, this comes at the cost of higher latency compared to the other optimization objectives.

\section{Consistency of Experimental Results}
\label{app:consistency}
Each optimization is repeated three times for fair evaluation. To assess consistency, we compute the Hypervolume (HV) indicator across runs with the same configuration. HV measures the portion of the objective space dominated by the Pareto front, relative to a reference point. Across all model-hardware optimizations, the standard deviation of HV remains below 0.03, with an average coefficient of variation (std/mean) under 3.5\%, indicating stable and reproducible results. Table~\ref{tab:consistency} summarizes these statistics for each model architecture and optimization objective.

\begin{table}[]
\caption{Summary of Hypervolume indicator difference across runs of the same configuration}
\label{tab:consistency}
\resizebox{1\columnwidth}{!}{
\begin{tabular}{ccccc}
\hline
\multicolumn{1}{c}{\textbf{Model}} & \multicolumn{1}{c}{\textbf{Metric}} & \multicolumn{1}{c}{\textbf{Hypervolume Mean}} & \multicolumn{1}{c}{\textbf{\begin{tabular}[c]{@{}c@{}}Standard Deviation\\ ($\sigma$)\end{tabular}}} & \multicolumn{1}{c}{\textbf{\begin{tabular}[c]{@{}c@{}}Coefficient of Variation\\ (\%)\end{tabular}}} \\ \hline
ViT-B/16 & carbon & 0.27 & 0.0006 & 0.2 \\ 
 & latency & 0.04 & 0.0001 & 0.3 \\
 & energy & 0.07 & 0.007 & 9.7 \\
 & carbon + latency & 0.26 & 0.009 & 3.5 \\ \hline
BertBase & carbon & 0.50 & 0.001 & 0.2 \\
 & latency & 0.70 & 0.004 & 0.6 \\
 & energy & 0.65 & 0.003 & 0.6 \\
 & carbon + latency & 0.48 & 0.013 & 2.8 \\ \hline
Llama3 & carbon & 0.45 & 0.033 & 7.4 \\
 & latency & 0.66 & 0.004 & 0.6 \\
 & energy & 0.35 & 0.003 & 1.1 \\
 & carbon + latency & 0.43 & 0.003 & 0.7 \\ \hline
CLIP ViT-B/16 & carbon & 0.19 & 0.021 & 11.1 \\
 & latency & 0.18 & 0.016 & 9.2 \\
 & energy & 0.39 & 0.015 & 3.8 \\
 & carbon + latency & 0.13 & 0.004 & 3.2 \\ \hline
\end{tabular}}
\end{table}

\section{CLIP ViT B/32 Accuracy Evaluations}
\label{app:b32}

In this section, we present results that demonstrate the applicability of \ourframework to the CLIP ViT-B/32 architecture and compare its performance against the TinyCLIP baselines. (Table~\ref{tab:b32}) Our findings show that \ourframework's optimization can be effectively generalized to other model architectures, yielding up to 5\% and 8\% reductions in carbon footprint while achieving higher accuracy and comparable latency compared to the TinyCLIP baselines, respectively.

\begin{scriptsize}
\begin{table*}[h]
\caption{The hardware and model architecture properties of each variant of the CarbonCLIP-B/32 family. Hardware configurations are specified as: $\{TC, PE_x, PE_y, L2, L2_{bw}, GLB \}$. Text and Vision encoders are specified as: \{Num Layers, FFN Dim, Hidden Dim, Num Heads\}}
\label{tab:b32}
\renewcommand{\arraystretch}{1.1} %
\resizebox{1\textwidth}{!}
{\begin{tabular}{lccccccc}
\hline
\multirow{2}{*}{Name} & \multirow{2}{*}{\begin{tabular}[c]{@{}c@{}}Carbon\\ (kgCO2e)\end{tabular}} & \multirow{2}{*}{\begin{tabular}[c]{@{}c@{}}Latency\\ (ms)\end{tabular}} & \multirow{2}{*}{\begin{tabular}[c]{@{}c@{}}Hardware\\ Configuration\end{tabular}} & \multicolumn{3}{c}{\begin{tabular}[c]{@{}c@{}}Model Configuration\end{tabular}} & \multirow{2}{*}{\begin{tabular}[c]{@{}c@{}}Avg. Accuracy \\ over 41 datasets\end{tabular}} \\ \cline{5-7}
 &  &  &  & \multicolumn{1}{c}{\begin{tabular}[c]{@{}c@{}}Text Encoder\\ Configuration\end{tabular}} & \multicolumn{1}{c}{\begin{tabular}[c]{@{}c@{}}Vision Encoder\\ Configuration\end{tabular}} & Params (M) &  \\ \hline

 CLIP-B/32 - DataComp & 0.42 & 15.1 & \{1, 32, 32 ,64 , 64,2\} & \multicolumn{1}{c}{\{12, 2048, 512, 8\}} & \multicolumn{1}{c}{\{12, 3072, 768, 12\}} & 144 & 51.1 \\ \hline
TinyCLIP-39M/32 & 0.37 & 3.0 & \{1, 32, 32, 64, 32, 2 \} & \multicolumn{1}{c}{\{6, 2048, 512, 8\}} & \multicolumn{1}{c}{\{12, 2048, 512, 8\}} & 84 & 45.2 \\
TinyCLIP-61M/32 & 0.39 & 9.4 & \{1, 128, 8, 64, 64, 2\} & \multicolumn{1}{c}{\{9, 2048, 512, 8\}} & \multicolumn{1}{c}{\{12, 2560, 640, 10\}} & 115 & 47.2 \\ \hline

\textbf{CarbonCLIP-32-S} & 0.35 & 7.3 & \{1, 64, 16, 64, 64, 2\} & \multicolumn{1}{c}{\{6, 1536, 384, 5\}} & \multicolumn{1}{c}{\{10, 3072, 672, 12\}} & 89 & 46.4 \\
\textbf{CarbonCLIP-32-M} & 0.36 & 15.3 & \{1, 64, 8, 64, 64, 2 \} & \multicolumn{1}{c}{\{8, 1280, 448, 7\}} & \multicolumn{1}{c}{\{11, 2688, 768, 8\}} & 99 & 47.6 \\
\textbf{CarbonCLIP-32-L} & 0.38 & 9.1 & \{1, 128, 8, 64, 128, 2\} & \multicolumn{1}{c}{\{7,2048,512,6\}} & \multicolumn{1}{c}{\{11,3072,768,10\}} & 113 & 49.1 \\ \hline
\end{tabular}}
\end{table*}
\end{scriptsize}

\section{CLIP Benchmark Full Result}
\label{app:carbonclip}
In this section we provide a detailed breakdown on the accuracy of each dataset for CarbonCLIP and the evaluated baselines in Table~\ref{tab:full_acc}.

\begin{table}[]
\caption{Results across all 41 evaluation benchmarks from CLIP Benchmark}

\label{tab:full_acc}
\renewcommand{\arraystretch}{1.1} %
\resizebox{\textwidth}{!}{%
\begin{tabular}{lcccccccccccccc}
\hline
\multicolumn{1}{c}{\multirow{2}{*}{\textbf{Dataset}}} & \multicolumn{5}{c}{\textbf{CarbonCLIP B/16 (ours)}} & \multicolumn{2}{c}{\textbf{TinyCLIP B/16}} & \textbf{DataComp B/16} & \multicolumn{3}{c}{\textbf{CarbonCLIP B/32 (ours)}} & \multicolumn{2}{c}{\textbf{TinyCLIP B/32}} & \textbf{DataComp B/32} \\ \cline{2-15} 
\multicolumn{1}{c}{} & XS & S & M & L & XL & 39M/16 & 8M/16 & ViT-B-16 & S & M & L & 40M/32 & 61M/32 & ViT-B-32 \\ \hline
cars & 0.74 & 0.82 & 0.84 & 0.85 & 0.87 & 0.52 & 0.08 & 0.89 & 0.82 & 0.83 & 0.84 & 0.77 & 0.8 & 0.87 \\
country211 & 0.11 & 0.14 & 0.16 & 0.17 & 0.2 & 0.18 & 0.12 & 0.22 & 0.15 & 0.16 & 0.17 & 0.13 & 0.15 & 0.18 \\
fer2013 & 0.17 & 0.21 & 0.3 & 0.36 & 0.34 & 0.52 & 0.33 & 0.39 & 0.25 & 0.23 & 0.38 & 0.47 & 0.49 & 0.33 \\
fgvc\_aircraft & 0.12 & 0.2 & 0.23 & 0.23 & 0.29 & 0.15 & 0.07 & 0.3 & 0.2 & 0.21 & 0.23 & 0.14 & 0.18 & 0.25 \\
flickr30k & 0.55 & 0.66 & 0.7 & 0.7 & 0.76 & 0.76 & 0.52 & 0.76 & 0.65 & 0.67 & 0.68 & 0.68 & 0.71 & 0.7 \\
flickr8k & 0.52 & 0.62 & 0.65 & 0.64 & 0.7 & 0.71 & 0.5 & 0.7 & 0.62 & 0.64 & 0.66 & 0.63 & 0.66 & 0.65 \\
gtsrb & 0.25 & 0.4 & 0.46 & 0.5 & 0.55 & 0.32 & 0.11 & 0.55 & 0.47 & 0.49 & 0.52 & 0.38 & 0.3 & 0.52 \\
imagenet-a & 0.11 & 0.19 & 0.26 & 0.32 & 0.39 & 0.33 & 0.15 & 0.48 & 0.22 & 0.22 & 0.24 & 0.17 & 0.21 & 0.3 \\
imagenet-o & 0.55 & 0.55 & 0.51 & 0.44 & 0.44 & 0.49 & 0.4 & 0.43 & 0.49 & 0.49 & 0.5 & 0.52 & 0.51 & 0.5 \\
imagenet-r & 0.55 & 0.66 & 0.73 & 0.77 & 0.82 & 0.7 & 0.3 & 0.84 & 0.72 & 0.74 & 0.75 & 0.7 & 0.73 & 0.78 \\
imagenet1k & 0.51 & 0.6 & 0.63 & 0.65 & 0.7 & 0.63 & 0.41 & 0.74 & 0.62 & 0.64 & 0.65 & 0.6 & 0.62 & 0.69 \\
imagenet\_sketch & 0.35 & 0.45 & 0.5 & 0.52 & 0.57 & 0.4 & 0.1 & 0.6 & 0.49 & 0.51 & 0.52 & 0.47 & 0.5 & 0.57 \\
imagenetv2 & 0.43 & 0.53 & 0.55 & 0.58 & 0.62 & 0.56 & 0.35 & 0.66 & 0.54 & 0.55 & 0.58 & 0.51 & 0.54 & 0.61 \\
mnist & 0.28 & 0.58 & 0.7 & 0.66 & 0.7 & 0.37 & 0.1 & 0.76 & 0.69 & 0.75 & 0.71 & 0.51 & 0.6 & 0.81 \\
mscoco\_captions & 0.33 & 0.41 & 0.44 & 0.44 & 0.49 & 0.47 & 0.29 & 0.49 & 0.41 & 0.43 & 0.44 & 0.41 & 0.45 & 0.45 \\
objectnet & 0.37 & 0.46 & 0.51 & 0.54 & 0.6 & 0.43 & 0.19 & 0.64 & 0.47 & 0.5 & 0.51 & 0.41 & 0.44 & 0.55 \\
renderedsst2 & 0.51 & 0.5 & 0.56 & 0.54 & 0.56 & 0.5 & 0.5 & 0.52 & 0.5 & 0.51 & 0.52 & 0.52 & 0.54 & 0.48 \\
stl10 & 0.92 & 0.94 & 0.96 & 0.97 & 0.98 & 0.97 & 0.92 & 0.98 & 0.96 & 0.96 & 0.96 & 0.95 & 0.96 & 0.97 \\
sun397 & 0.59 & 0.66 & 0.68 & 0.69 & 0.71 & 0.69 & 0.56 & 0.71 & 0.67 & 0.67 & 0.68 & 0.65 & 0.67 & 0.68 \\
voc2007 & 0.7 & 0.73 & 0.75 & 0.77 & 0.79 & 0.77 & 0.62 & 0.82 & 0.77 & 0.77 & 0.77 & 0.77 & 0.78 & 0.81 \\
voc2007\_multilabel & 0.75 & 0.79 & 0.81 & 0.81 & 0.83 & 0.82 & 0.74 & 0.81 & 0.79 & 0.8 & 0.8 & 0.76 & 0.79 & 0.79 \\
vtab/caltech101 & 0.8 & 0.83 & 0.84 & 0.84 & 0.85 & 0.82 & 0.72 & 0.85 & 0.83 & 0.83 & 0.85 & 0.82 & 0.82 & 0.84 \\
vtab/cifar10 & 0.73 & 0.83 & 0.89 & 0.9 & 0.93 & 0.91 & 0.73 & 0.96 & 0.91 & 0.92 & 0.92 & 0.91 & 0.92 & 0.96 \\
vtab/cifar100 & 0.47 & 0.55 & 0.65 & 0.67 & 0.75 & 0.68 & 0.42 & 0.82 & 0.69 & 0.7 & 0.73 & 0.69 & 0.72 & 0.8 \\
vtab/clevr\_closest\_object\_distance & 0.16 & 0.15 & 0.17 & 0.16 & 0.19 & 0.2 & 0.16 & 0.24 & 0.16 & 0.16 & 0.16 & 0.17 & 0.21 & 0.21 \\
vtab/clevr\_count\_all & 0.18 & 0.19 & 0.21 & 0.34 & 0.25 & 0.2 & 0.13 & 0.33 & 0.16 & 0.2 & 0.33 & 0.19 & 0.24 & 0.13 \\
vtab/diabetic\_retinopathy & 0.04 & 0.09 & 0.16 & 0.04 & 0.2 & 0.03 & 0.02 & 0.11 & 0.05 & 0.07 & 0.05 & 0.1 & 0.24 & 0.42 \\
vtab/dmlab & 0.14 & 0.15 & 0.15 & 0.15 & 0.14 & 0.13 & 0.18 & 0.19 & 0.21 & 0.2 & 0.14 & 0.21 & 0.15 & 0.16 \\
vtab/dsprites\_label\_orientation & 0.02 & 0.02 & 0.02 & 0.02 & 0.02 & 0.02 & 0.03 & 0.02 & 0.02 & 0.04 & 0.03 & 0.02 & 0.02 & 0.03 \\
vtab/dsprites\_label\_x\_position & 0.03 & 0.03 & 0.03 & 0.03 & 0.03 & 0.03 & 0.03 & 0.03 & 0.03 & 0.02 & 0.03 & 0.03 & 0.03 & 0.03 \\
vtab/dsprites\_label\_y\_position & 0.03 & 0.03 & 0.03 & 0.03 & 0.03 & 0.03 & 0.04 & 0.03 & 0.03 & 0.03 & 0.03 & 0.03 & 0.03 & 0.03 \\
vtab/dtd & 0.41 & 0.51 & 0.54 & 0.51 & 0.58 & 0.47 & 0.29 & 0.58 & 0.48 & 0.51 & 0.53 & 0.51 & 0.52 & 0.57 \\
vtab/eurosat & 0.39 & 0.48 & 0.54 & 0.59 & 0.58 & 0.53 & 0.23 & 0.59 & 0.5 & 0.5 & 0.56 & 0.48 & 0.45 & 0.57 \\
vtab/flowers & 0.55 & 0.66 & 0.64 & 0.66 & 0.71 & 0.7 & 0.58 & 0.76 & 0.63 & 0.66 & 0.68 & 0.62 & 0.64 & 0.73 \\
vtab/kitti\_closest\_vehicle\_distance & 0.37 & 0.32 & 0.3 & 0.26 & 0.35 & 0.11 & 0.15 & 0.29 & 0.28 & 0.19 & 0.32 & 0.15 & 0.17 & 0.16 \\
vtab/pcam & 0.59 & 0.57 & 0.59 & 0.63 & 0.6 & 0.61 & 0.53 & 0.6 & 0.61 & 0.58 & 0.54 & 0.52 & 0.57 & 0.53 \\
vtab/pets & 0.76 & 0.85 & 0.87 & 0.89 & 0.91 & 0.81 & 0.46 & 0.93 & 0.87 & 0.88 & 0.89 & 0.85 & 0.88 & 0.9 \\
vtab/resisc45 & 0.45 & 0.57 & 0.62 & 0.64 & 0.66 & 0.55 & 0.21 & 0.65 & 0.58 & 0.64 & 0.63 & 0.54 & 0.58 & 0.63 \\
vtab/smallnorb\_label\_azimuth & 0.05 & 0.06 & 0.06 & 0.05 & 0.05 & 0.06 & 0.06 & 0.05 & 0.05 & 0.05 & 0.05 & 0.05 & 0.05 & 0.05 \\
vtab/smallnorb\_label\_elevation & 0.11 & 0.11 & 0.12 & 0.11 & 0.11 & 0.1 & 0.12 & 0.11 & 0.11 & 0.1 & 0.11 & 0.11 & 0.11 & 0.1 \\
vtab/svhn & 0.15 & 0.3 & 0.29 & 0.29 & 0.38 & 0.16 & 0.14 & 0.61 & 0.33 & 0.45 & 0.42 & 0.35 & 0.39 & 0.61 \\ \hline
\end{tabular}%
}
\end{table}

\section{Carbon Footprint Breakdown}
\label{app:breakdown}
\begin{figure}
\centering
\includegraphics[width=0.6\columnwidth]{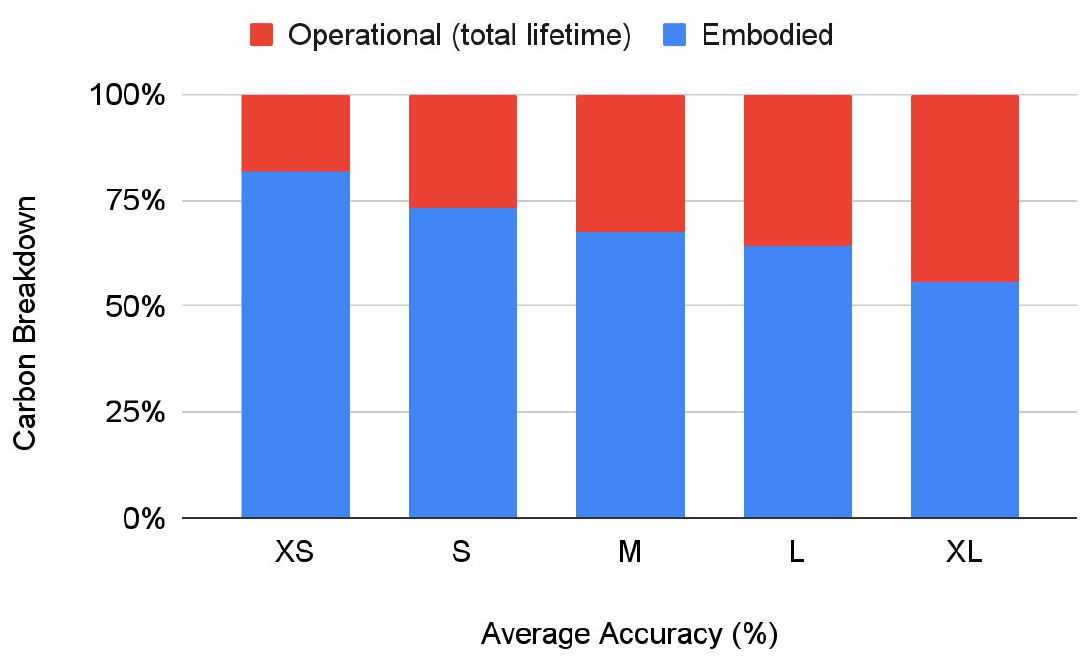}
\caption{Operational and embodied carbon footprint breakdown for the CLIP-ViT-B/16 architecture.}
\label{fig:carbon_breakdown}
\vspace{-3ex}
\end{figure}

We provide a breakdown of the carbon footprint for each variant of the CarbonCLIP family model. As shown in Figure~\ref{fig:carbon_breakdown}, as the model increases in size, the proportion of operational carbon in the overall carbon footprint of the model increases from 20\% to over 40\%. The CarbonCLIP-XL model has 3 $\times$ the number of parameters and almost 3 $\times$ the latency of CarbonCLIP-XS, but the selected hardware architecture only has double the number of compute PEs. Therefore, the operational carbon increases proportionally more than the increase in embodied carbon. This highlights need of co-optimizing the model and hardware architecture to maintain an intricate balance between operational and embodied carbon, keeping the overall carbon footprint of the system low. As such, the expected lifetime and source of power are also important factors that need to be taken into account during the optimization process.

\section{Case Study: Varying Latency Constraints}
\label{app:latency}

\begin{figure}
\centering
\includegraphics[width=0.6\columnwidth]{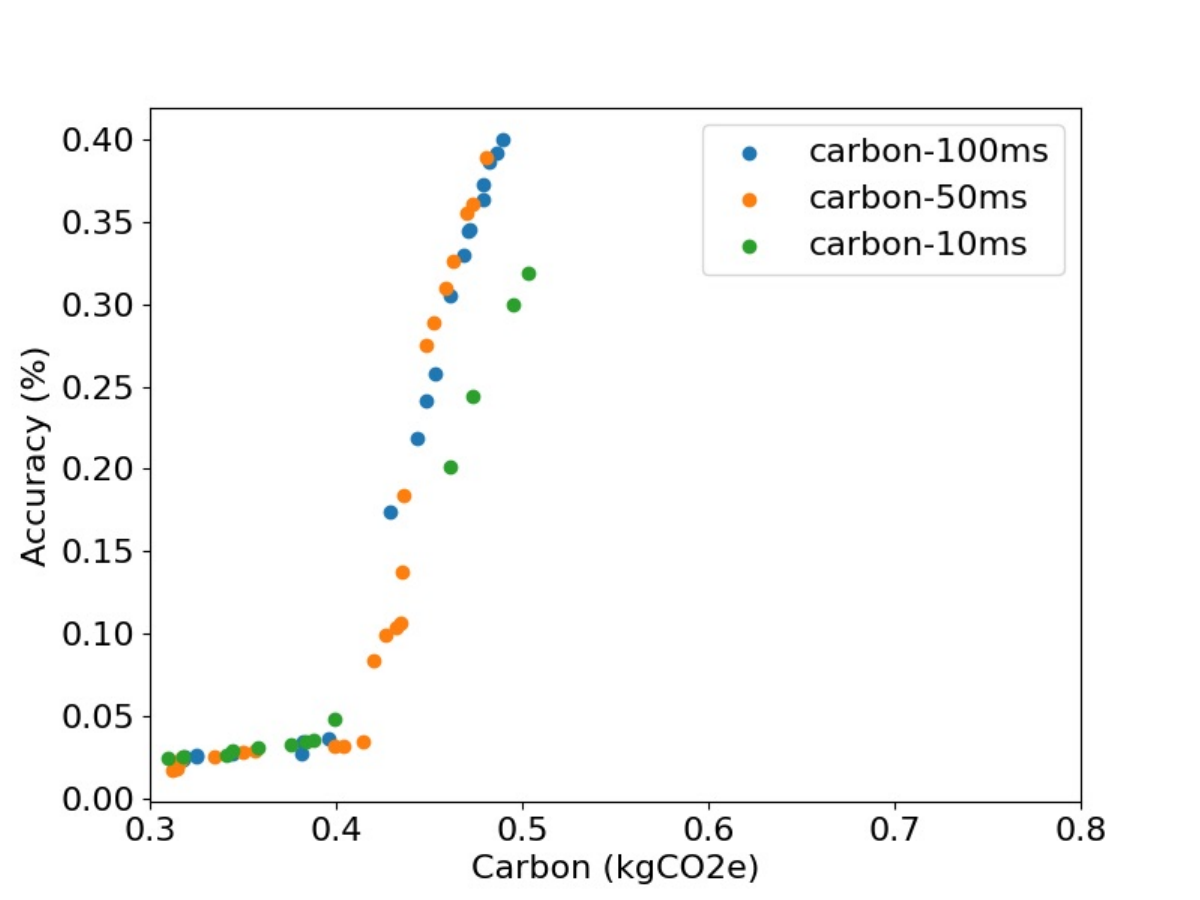}
\caption{Latency and carbon footprint trade-offs for the CLIP-ViT-B/16 architecture across the Pareto frontier.}
\label{fig:latency}
\vspace{-3ex}
\end{figure}

We evaluate the impact of latency constraints on model and hardware configurations, as well as their carbon footprints. We categorize use cases into three categories based on latency requirements: critical real-time ($<$10ms), interactive ($<$50ms), and non-critical ($<$100ms)~\cite{latency}. Figure~\ref{fig:latency} shows the Pareto frontiers for each category.

Lower latency constraints typically result in less carbon-efficient designs. For example, a 10ms latency constraint achieves only an 17\% carbon reduction compared to latency-optimized models, although with comparable latency values. However, increasing the constraint to 100ms does not consistently improve efficiency, as many optimal designs already meet the 50ms threshold.

\section{Case Study: Varying Operational Regions}
\label{app:region}
\begin{figure}
\centering
\includegraphics[width=0.45\columnwidth]{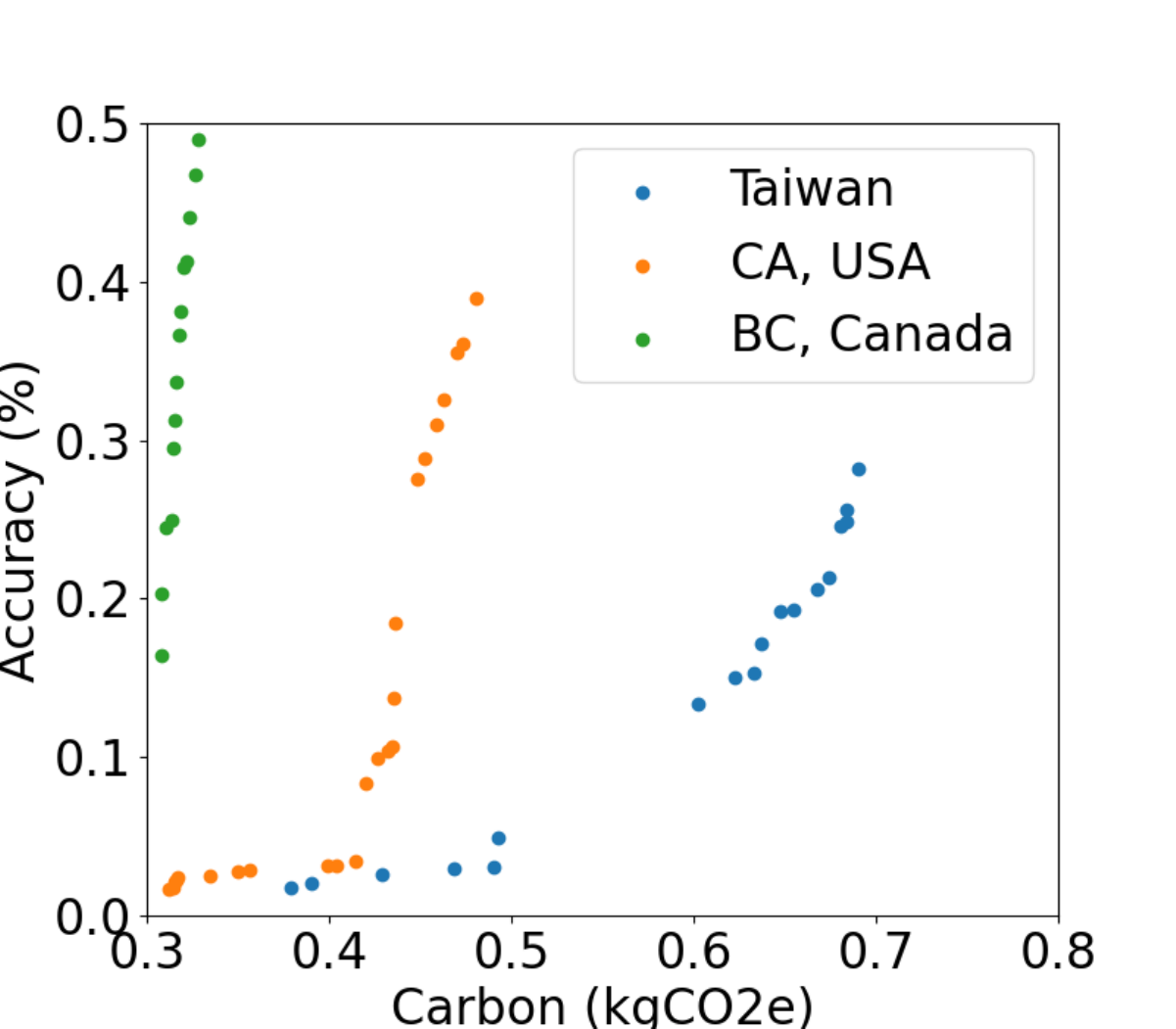}
\caption{Pareto Frontier of Carbon-only Optimization searches across High (Taiwan), Mid (California, USA), and Low (British Columbia, Canada) carbon intensity regions.}
\label{fig:op_region}
\vspace{-3ex}
\end{figure}

\begin{figure}[t!]
    \centering
    \begin{subfigure}[t]{0.45\textwidth}
        \centering
        \includegraphics[width=\textwidth]{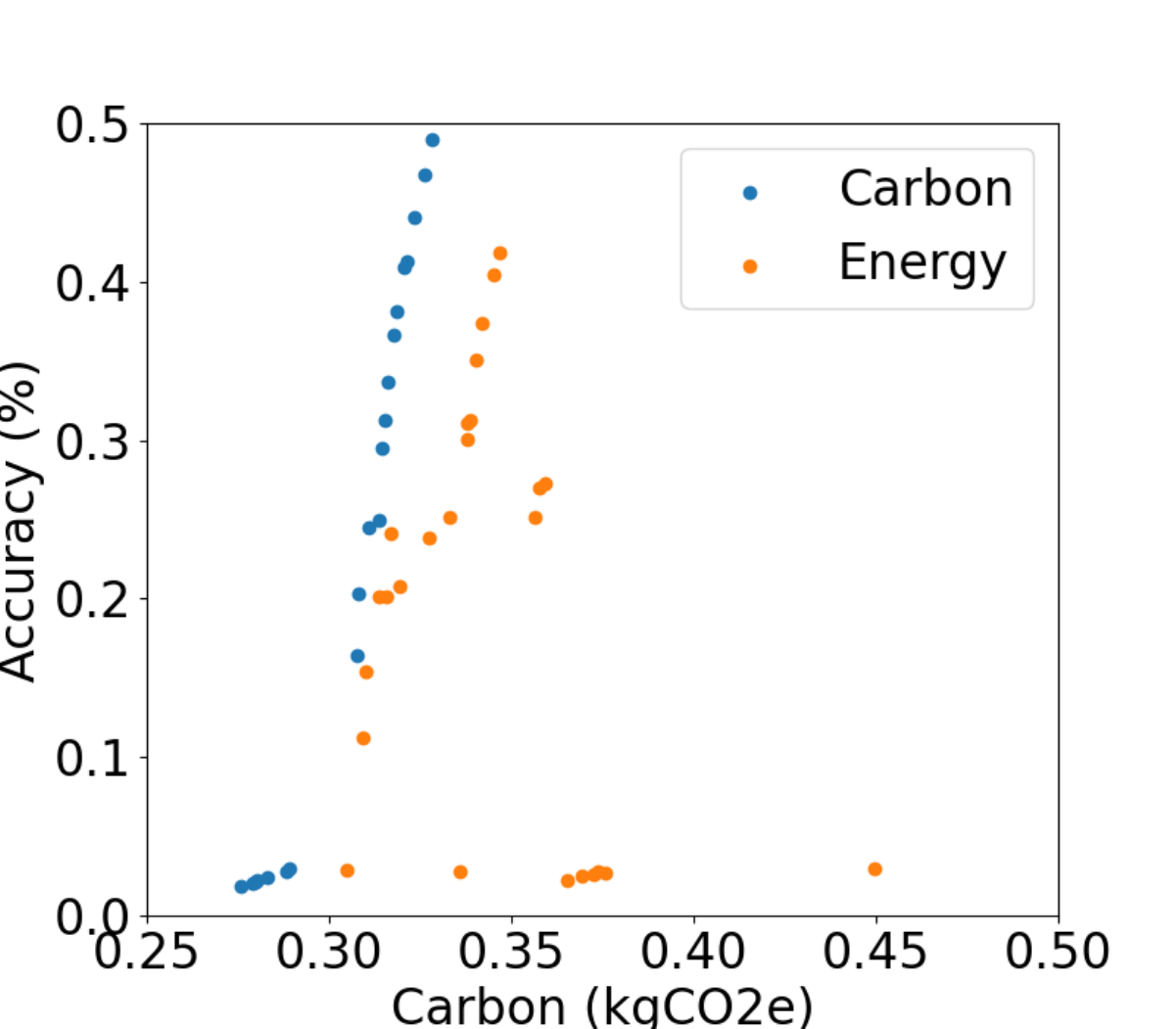}
        \caption{British Columbia (Low carbon intensity)}
        \label{fig:bc}
    \end{subfigure}
    \hfill
    \begin{subfigure}[t]{0.45\textwidth}
        \centering
        \includegraphics[ width=\textwidth]{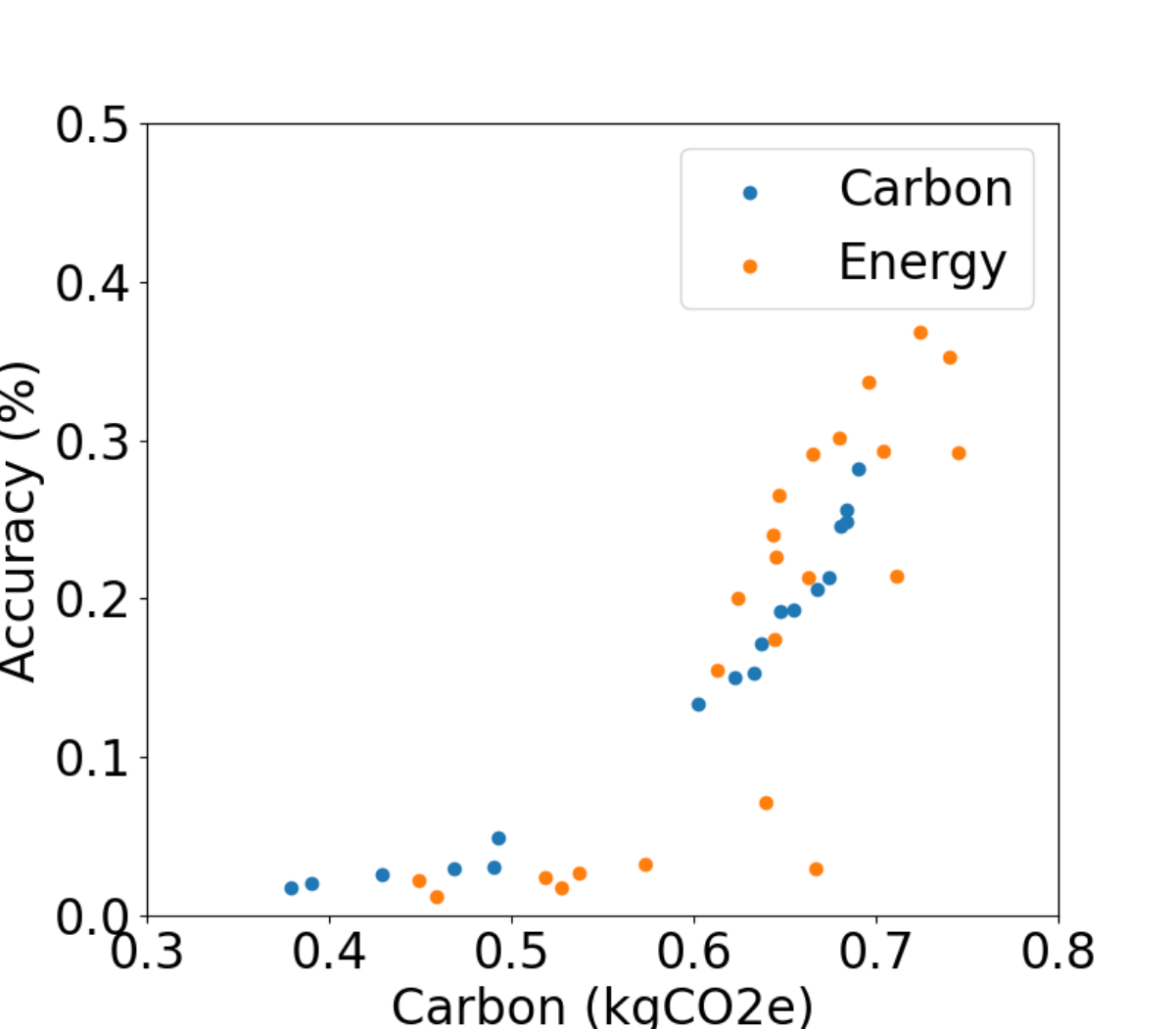}
        \caption{Taiwan (High carbon intensity)}
        \label{fig:tw}
    \end{subfigure}
    \caption{Pareto Frontier of the Carbon vs Energy Optimization result in regions with different carbon intensity.}
\end{figure}

The region of the operation and manufacturing of the model affects the search results, and the quality of the metric immensely. Figure~\ref{fig:op_region} shows that varying the operational carbon region, in high, mid, low carbon-intensity regions, yields dramatically different results in terms of the model searched and resulting configurations. Additionally, while our case studies (Section~\ref{sec:compute_constraints}) show that energy optimization is effective when operational carbon dominates (e.g., smaller architectures, high-carbon-intensity regions), in cleaner energy regions, total carbon optimization still yields lower emissions. Optimizing against total carbon instead of energy can reduce emissions by 8\% in low-carbon regions (Canada) vs. 2\% in high-carbon regions (Taiwan). The pareto frontier of each region is shown in Figure~\ref{fig:bc} and Figure~\ref{fig:tw}.

\begin{figure}[h]
    \centering
    \begin{subfigure}{0.47\textwidth}
        \centering
        \includegraphics[width=\textwidth]{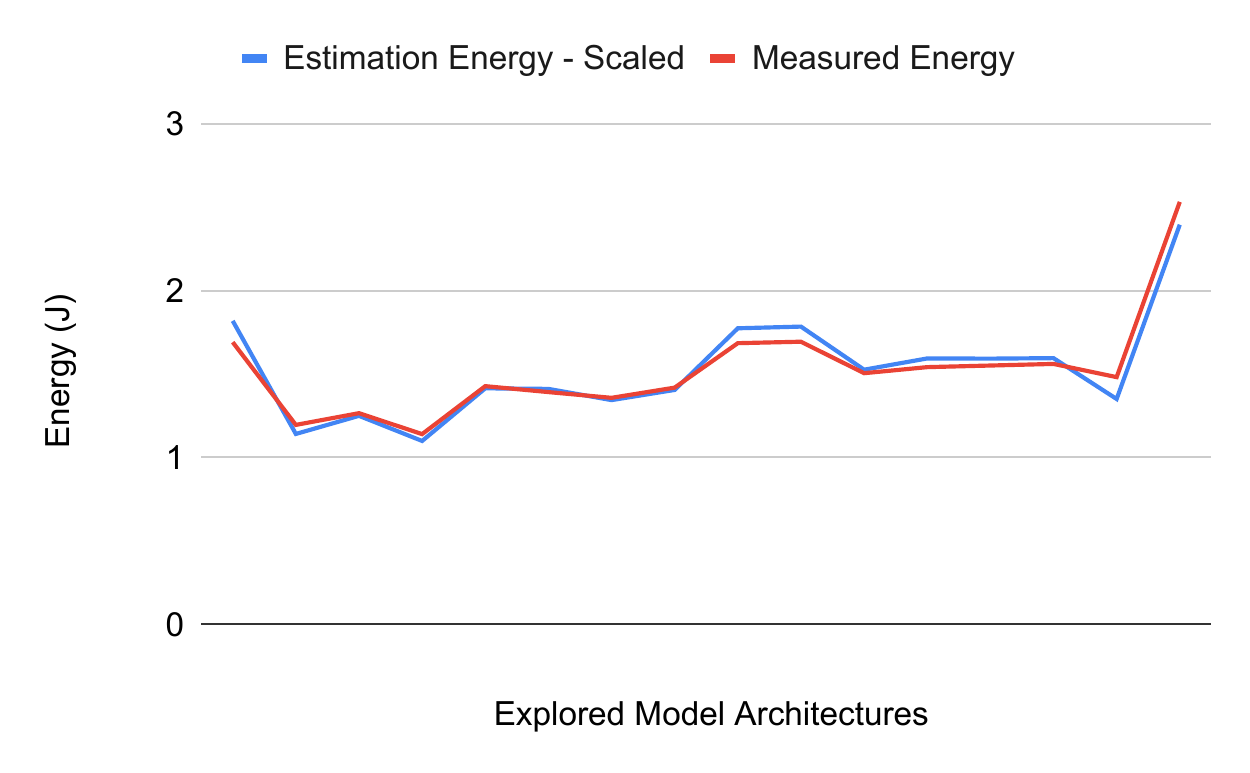}
        \caption{V100 - Energy}

    \end{subfigure}
    \hfill
    \begin{subfigure}{0.48\textwidth}
        \centering
        \includegraphics[ width=\textwidth]{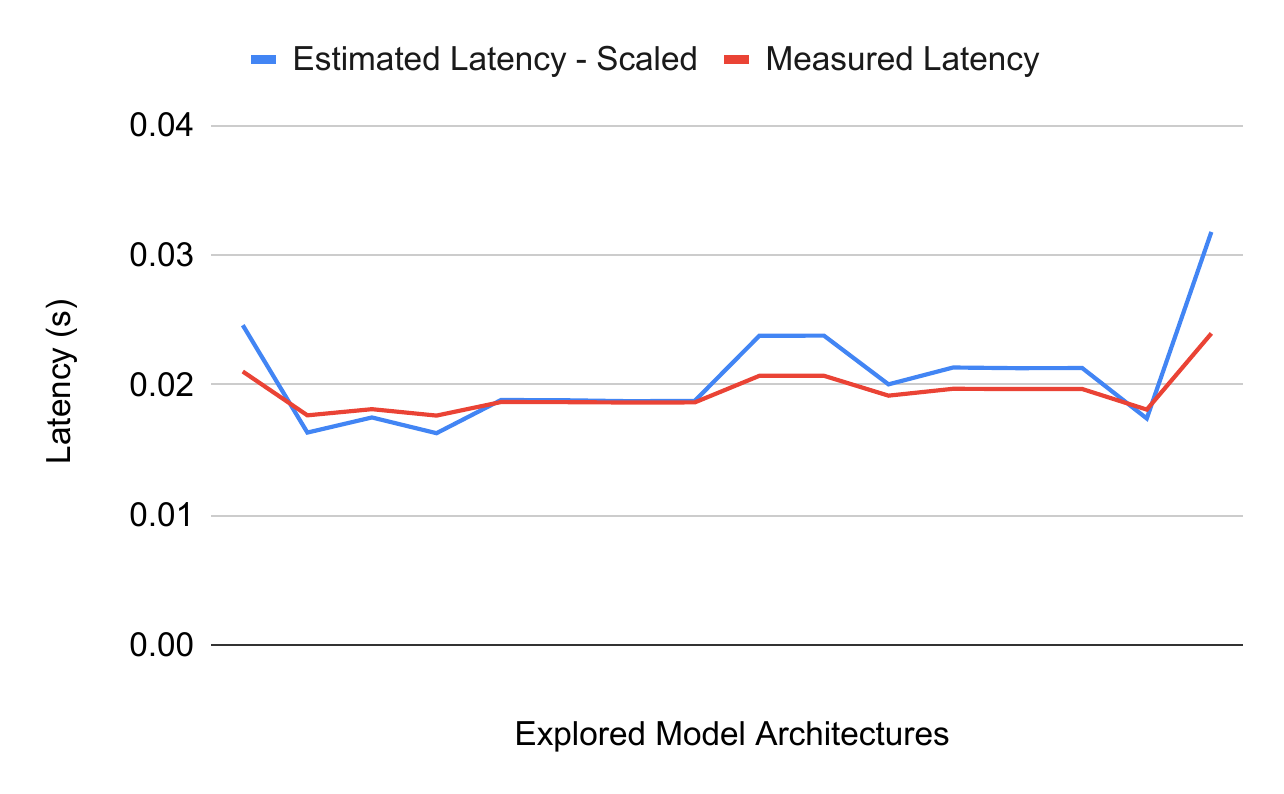}
        \caption{V100 - Latency}

    \end{subfigure}
    \begin{subfigure}{0.47\textwidth}
        \centering
        \includegraphics[width=\textwidth]{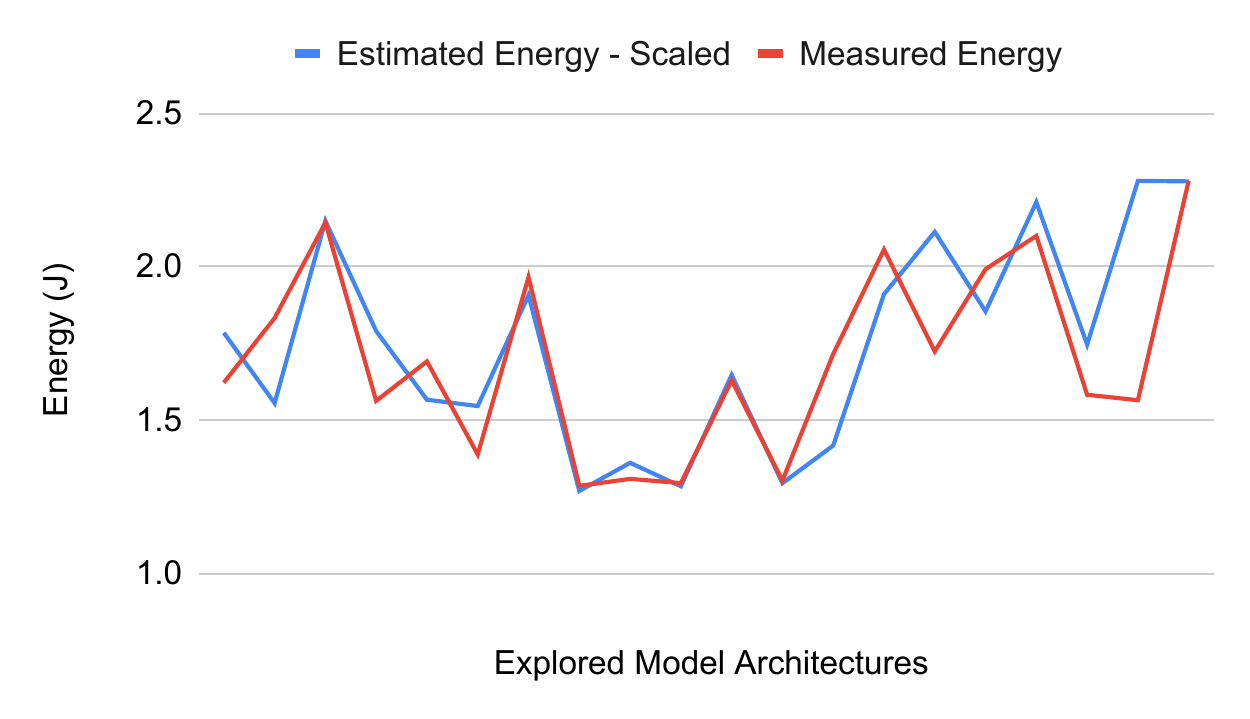}
        \caption{A100 - Energy}

    \end{subfigure}
    \hfill
    \begin{subfigure}{0.48\textwidth}
        \centering
        \includegraphics[ width=\textwidth]{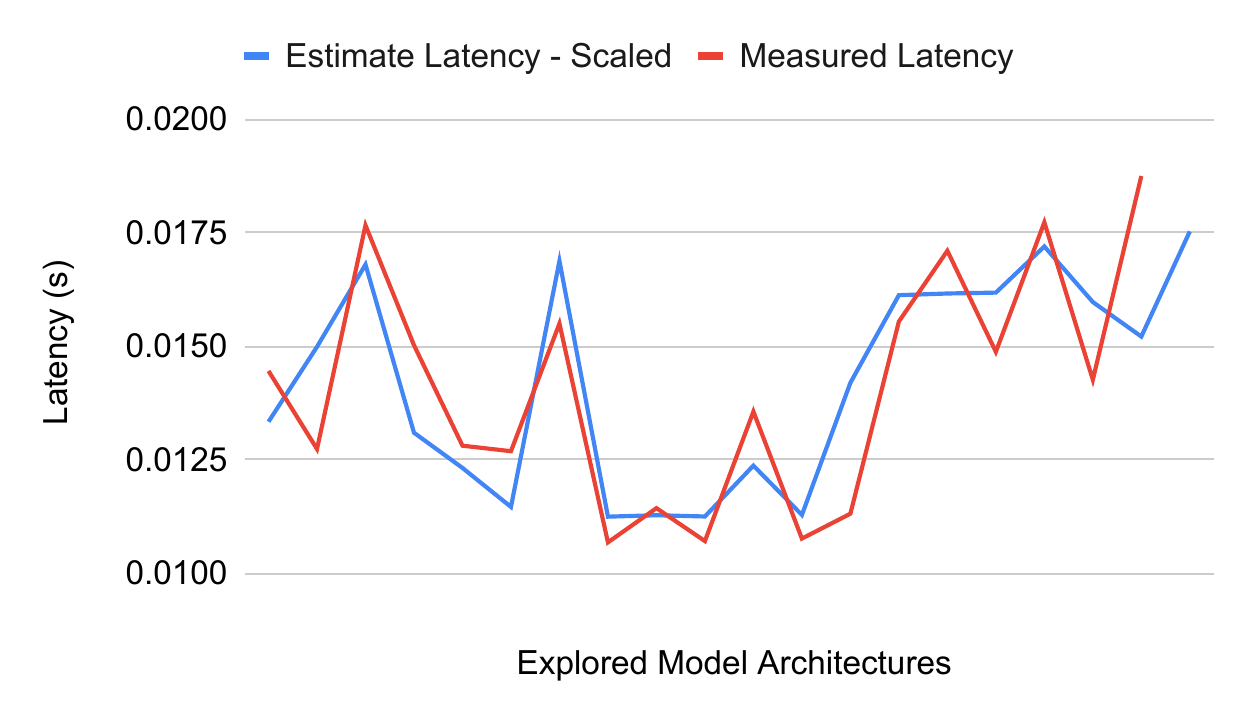}
        \caption{A100 - Latency}

    \end{subfigure}
    \begin{subfigure}{0.47\textwidth}
        \centering
        \includegraphics[width=\textwidth]{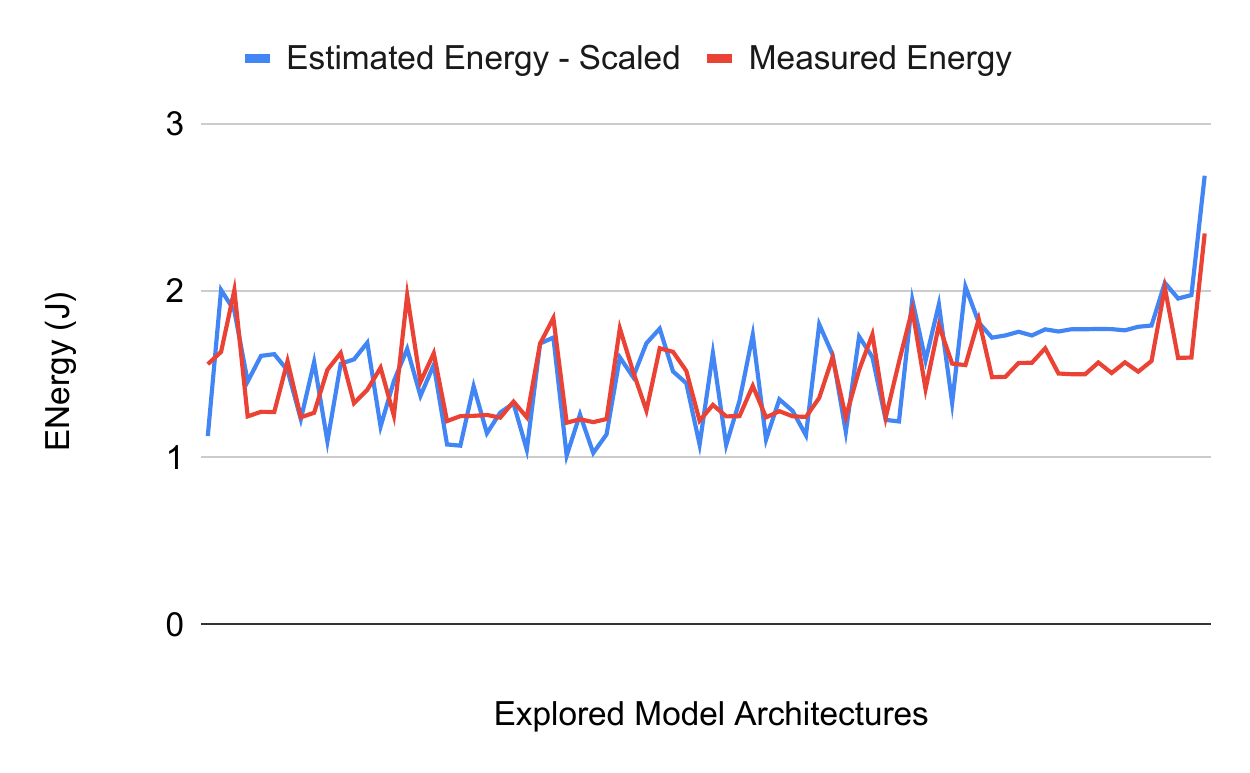}
        \caption{H100 - Energy}

    \end{subfigure}
    \hfill
    \begin{subfigure}{0.48\textwidth}
        \centering
        \includegraphics[ width=\textwidth]{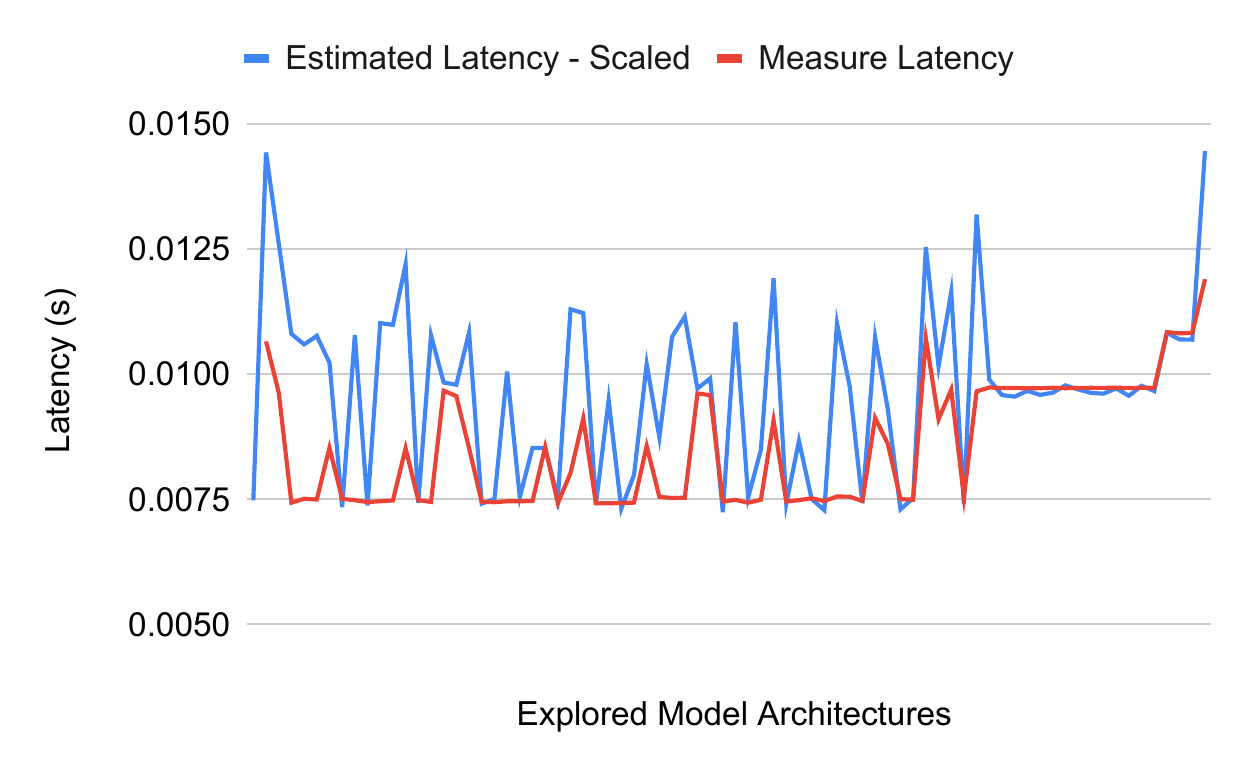}
        \caption{H100 - Latency}

    \end{subfigure}
    \caption{GPU estimated vs measure performance.}
    \label{fig:gpu_validation}
\end{figure}

\section{Case Study: Model Architecture Search with GPUs}
\label{app:model_gpu}

 Once a hardware architecture is selected, \ourframework ~supports model architecture search on fixed hardware to refine configurations and re-evaluate carbon costs before manufacturing.

Table~\ref{tab:fixed_arch} shows an example of model architecture search on a fixed V100-like hardware architecture, optimizing total carbon and latency.

\begin{table}[]
\caption{Model architecture search with fixed V100-like architecture. Text and Vision encoders are specified in the format of \{Num Layers, FFN Dim, Hidden Dim, Num Heads\}}
\label{tab:fixed_arch}
\resizebox{\textwidth}{!}{%
\begin{tabular}{ccccc}
\hline
Proxy Accuracy (\%) & Carbon (kgCO2e) & Latency (ms) & Text Model Config & Vision Model Config \\ \hline
2.8 & 11.13 & 37 & 6,1024,512,6 & 6,1536,672,6 \\ 
9.8 & 11.41 & 58 & 6,2048,512,8 & 9,3072,768,6 \\ 
14.4 & 11.45 & 64 & 6,2048,512,6 & 10,2688,768,6 \\ 
25.3 & 11.57 & 76 & 6,1024,512,7 & 12,2688,768,6 \\ 
28.7 & 11.58 & 76 & 6,1792,512,7 & 12,2688,768,6 \\ \hline
\end{tabular}%
}
\end{table}

\section{Energy and Latency Validation with GPUs}
\label{app:gpu_validation}

To maintain high confidence in our estimation toolchains, we validated our energy and latency estimates using existing GPU hardware (V100, A100, and H100). Although GPUs differ from domain-specific accelerators, we modeled a GPU-like architecture with comparable tensor and vector units, conducted a model search on the fixed architecture, and profiled actual hardware performance for each searched data point.

The results demonstrate a strong correlation between our estimates and measured performance, with Spearman’s rank-order correlation ranging from 0.5 to 1.0. When energy and latency values are scaled to a common measurement range using constants, the estimates yield average errors of 8\% for energy and 9\% for latency across GPU architectures. This validation confirms that our estimation tools provide accurate and reliable results, even with simulated hardware. Table~\ref{tab:gpu_validation} summarizes the results, while Figure~\ref{fig:gpu_validation} presents the estimated and measured performance for each evaluation GPU architecture. 

\begin{table}[]
\centering
\caption{Latency and Energy estimation error against Real GPU hardware}
\label{tab:gpu_validation}
\renewcommand{\arraystretch}{1.1} %
\resizebox{\textwidth}{!}{%
\begin{tabular}{ccccc}
\hline
\textbf{Hardware} & \textbf{Energy Estimation Error} & \textbf{Spearman's $r$ (Energy)} &\textbf{Latency Estimation Error} &\textbf{Spearman's $r$ (Latency)} \\ \hline
V100 & 3.5\% & 0.97 & 7.4\% & 1.0\\ 
A100 & 8.4\% & 0.74 & 8.5\% & 0.85\\ 
H100 & 12.3\% & 0.7 &11.1\% & 0.5\\ \hline
\end{tabular}%
}
\end{table}

\section{Carbon Footprint of the Framework}
\label{app:carbonFramework}
We quantify the carbon footprint of running \ourframework\ using CodeCarbon~\cite{codecarbon}. On average, the optimization process takes 5 hours for BERT\textsubscript{base} and up to 20 hours for CLIP models. For CLIP, the most resource-intensive case, 100 optimization trials emit approximately 57 kgCO\textsubscript{2}e, while final model training emits 454 kgCO\textsubscript{2}e per model. This means the optimization process costs roughly 1/13th the carbon budget of training the final model. Despite the one-time cost of optimization, CATransformer achieves overall efficiency gains through reduced training steps post-pruning, along with inference gains that scale with the number of devices.


\end{document}